\documentclass[10pt,twocolumn,letterpaper]{article}

\usepackage{cvpr}              

\newcommand{\red}[1]{{\color{red}#1}}

\definecolor{cvprblue}{rgb}{0.21,0.49,0.74}
\usepackage[pagebackref,breaklinks,colorlinks,allcolors=cvprblue]{hyperref}
\usepackage{pifont}
\usepackage{multirow}
\usepackage{adjustbox}
\usepackage{newtxmath}
\usepackage{bbm, dsfont}
\usepackage{xcolor}
\usepackage{colortbl}
\usepackage{dsfont}

\newcommand{\blue}[1]{\textcolor{blue}{#1}} 

\def\paperID{463} 
\def\confName{CVPR}
\def\confYear{2025}

\title{Luminance-GS: Adapting 3D Gaussian Splatting to Challenging Lighting Conditions with View-Adaptive Curve Adjustment}

\author{Ziteng Cui\textsuperscript{1},
    Xuangeng Chu\textsuperscript{1},
    Tatsuya Harada\textsuperscript{1,2}\\
    \textsuperscript{1} The University of Tokyo \textsuperscript{2} RIKEN AIP \\
    \small{(cui, xuangeng.chu, harada)@mi.t.u-tokyo.ac.jp}
    }

\begin{document}
\maketitle

\begin{abstract}

Capturing high-quality photographs under diverse real-world lighting conditions is challenging, as both natural lighting (e.g., low-light) and camera exposure settings (e.g., exposure time) significantly impact image quality. This challenge becomes more pronounced in multi-view scenarios, where variations in lighting and image signal processor (ISP) settings across viewpoints introduce photometric inconsistencies. Such lighting degradations and view-dependent variations pose substantial challenges to novel view synthesis (NVS) frameworks based on Neural Radiance Fields (NeRF) and 3D Gaussian Splatting (3DGS).

To address this, we introduce \textbf{Luminance-GS}, a novel approach to achieving high-quality novel view synthesis results under diverse challenging lighting conditions using 3DGS. By adopting per-view color matrix mapping and view adaptive curve adjustments, Luminance-GS achieves state-of-the-art (SOTA) results across various lighting conditions—including low-light, overexposure, and varying exposure—while not altering the original 3DGS explicit representation. Compared to previous NeRF- and 3DGS-based baselines, Luminance-GS provides real-time rendering speed with improved reconstruction quality. The source code is available at \footnote{\url{https://github.com/cuiziteng/Luminance-GS}}.
\end{abstract}

\section{Introduction}
\label{sec:intro}

\begin{figure}
    \centering
    \includegraphics[width=1.00\linewidth]{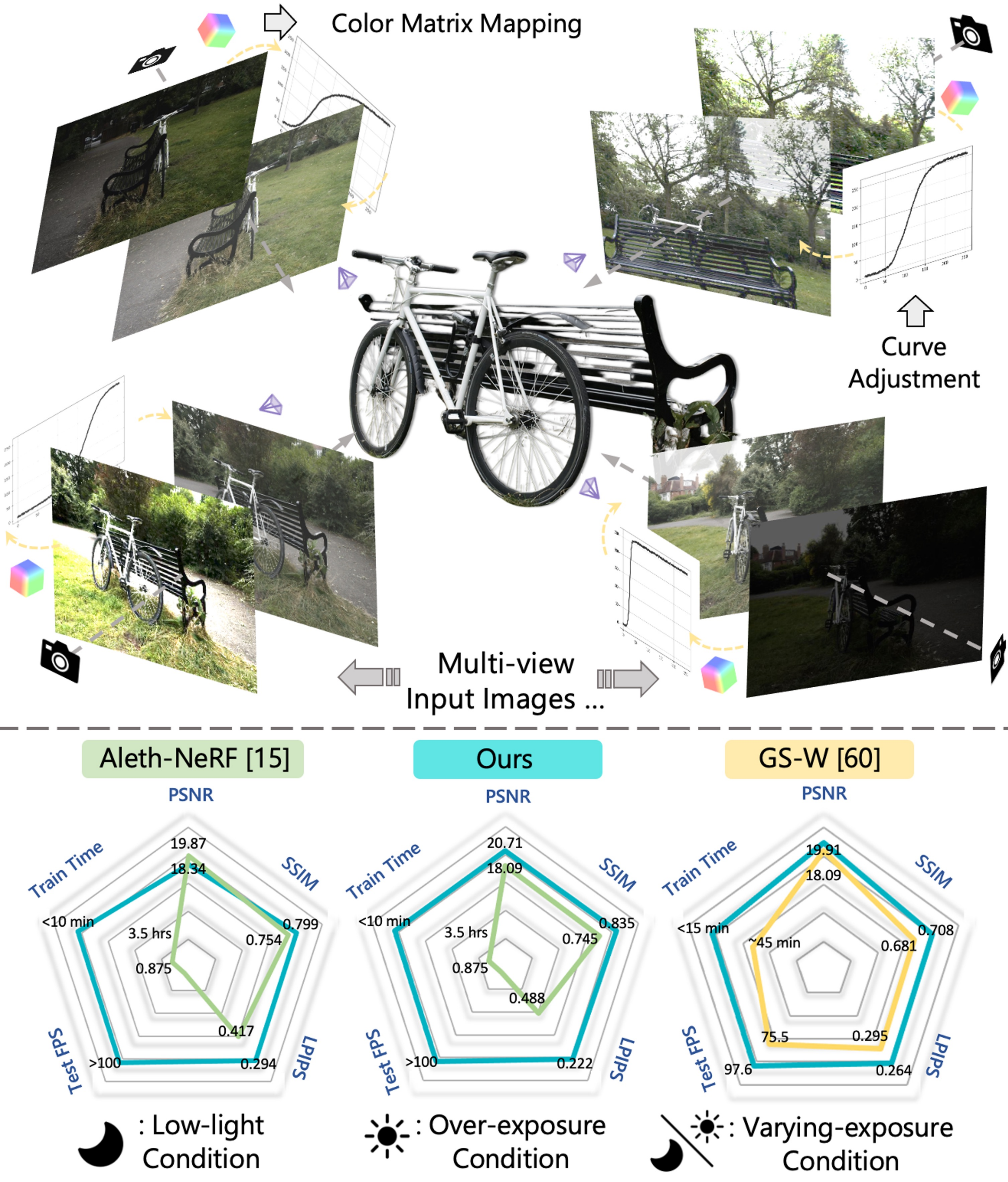}
    
    \caption{\textbf{\textit{Up}}: The core idea of Luminance-GS is to perform view-adaptive adjustments for images from each viewpoint, including color matrix mapping and curve adjustment. \textbf{\textit{Down}}: Compare with previous SOTA solutions~\cite{cui_aleth_nerf,GS-W_ECCV2024}, our approach achieves superior performance and efficiency across different lighting conditions.}
    \label{fig:motivation}
    \vspace{-4mm}
\end{figure}

In recent years, Neural Radiance Field (NeRF) and Gaussian Splatting (3DGS) based methods~\cite{nerf,NeRF++,Li_2023_ICCV,depth_nerf,barron2023zipnerf,barron2022mipnerf360,li2024nerfxl,3dgs,charatan23pixelsplat,Niedermayr_2024_CVPR,liu2025mvsgaussian} have made significant strides in reconstructing realistic 3D scenes and generating coherent novel views.  The impressive performance of NeRF and 3DGS has found applications across various fields, including augmented and virtual reality (AR/VR), computational photography, medical imaging, and autonomous driving, etc. Both NeRF- and 3DGS-based methods aim to synthesize novel views by learning from existing multi-view images, capturing consistent geometric information and simulating non-Lambertian effects. Despite their success under normal, uniform lighting, these methods often struggle with real-world challenging lighting conditions (e.g., low-light, varying camera exposure), which is primarily due to the lightness distortion caused by the environment or camera sensor limitations~\cite{raw_nerf,TOG_wang2024bilateral}, and the inherent lack of illumination modeling in both NeRF and 3DGS methods~\cite{rudnev2022nerfosr,CVPR24_GS_IR}.
To make matters worse, \textit{de facto} results~\cite{LLNeRF,cui_aleth_nerf,zou2024enhancing_aaai} also show that using 2D image restoration methods~\cite{Zero-DCE,cvpr22_sci,Cui_2022_BMVC,Afifi_2021_CVPR} to pre-process multi-view training images is also ineffective, as 2D models often fail to preserve 3D consistency, leading to floaters and artifacts during rendering.


 To ensure high-quality view synthesis under challenging lighting conditions, NeRF-W~\cite{martinbrualla2020nerfw} and its following works~\cite{HA-NeRF,yang2023CRNeRF,GS-W_ECCV2024,kulhanek2024wildgaussians} typically learn a separate appearance embedding for each viewpoint, effectively isolating the lighting information specific to each camera view. However, they are limited to capturing only the lighting conditions present in input images, restricting the models' ability to adapt to a new lighting condition during rendering (e.g., achieving low-light enhancement).
 For low-light conditions, NeRF-based solutions~\cite{LLNeRF,cui_aleth_nerf,zou2024enhancing_aaai,TOG_wang2024bilateral} tend to incorporate different assumptions in volume rendering stage, enabling unsupervised lightness restoration while maintaining 3D multi-view consistency. For instance, LL-NeRF~\cite{LLNeRF} decomposes NeRF's color MLP to separate each volume particle into view-dependent and view-independent components for distinct processing, meanwhile Aleth-NeRF~\cite{cui_aleth_nerf} introduces an additional MLP branch to learn ``concealing fields'', simulating low-light conditions as occlusions, which are later removed during testing to enhance output.

However, these solutions do not generalize well to explicit representations such as 3DGS~\cite{3dgs}, which employs rasterization to store the color of geometric primitives in an explicit radiance field, removing the need of implicit representation and MLP structure, thus offers substantial advantages over NeRF-based methods in both inference speed and training time~\cite{3dgs,charatan23pixelsplat,HDR_GS_nips,fan2023lightgaussian}. This raises a key question: How can we leverage the efficiency of 3DGS's explicit rendering while also ensuring robust reconstruction quality under diverse and challenging lighting conditions?


Our solution, \textbf{Luminance-GS}, avoids altering the original 3DGS explicit representation or imposing additional lighting assumptions. Instead, we employ a simple yet effective image processing technique: curve adjustment. We use different tone curves to map input images under varying lighting conditions (e.g., low-light, overexposure) to a consistently well-lit output, ensuring view consistency and supporting 3DGS training. To achieve this, we design two sub-solutions: per-view color matrix mapping and view-adaptive curve adjustment. For each viewpoint, we project the image into a suitable representation using a specific matrix, followed by curve adjustment tailored to the current view’s lighting condition (see Fig.~\ref{fig:motivation} \textbf{\textit{Up}}). During training time,  the projection matrix and view-adaptive curve would be jointly optimized  with other 3DGS parameters. Finally, several unsupervised loss functions are introduced to maintain the shape of the curve while ensuring view alignment across images from different viewpoints. Extensive experiments under various lighting conditions demonstrate our algorithm’s state-of-the-art (SOTA) performance, achieving significant advantages over previous methods in both efficiency and quality (see Fig.~\ref{fig:motivation} \textbf{\textit{Down}}). The contributions of our work can be summarized as follows:

\begin{itemize}
     \item We introduce Luminance-GS, a novel framework that extends 3DGS to handle novel view synthesis under diverse challenging lighting conditions. By using per-view color matrix mapping and view-adaptive curve adjustment, we achieve view-aligned, normal-light outputs from images captured under varying lighting conditions.
     
     \item We introduce unsupervised loss functions to guide both curve mapping and 3DGS training. Additionally, curve mapping is applied only during training to generate pseudo-enhanced images, saving a significant amount of testing time.
     
     \item Our method achieves state-of-the-art (SOTA) results in various lighting conditions (low-light, overexposure, and varying exposure), substantially improving both speed and rendering quality compared to previous methods. 
\end{itemize}

\section{Related Works}
\label{sec:related}

\subsection{3D Gaussian Splatting}
Neural Radiance Fields (NeRF)~\cite{nerf} have greatly advanced the field of novel view synthesis (NVS), but NeRF-based methods~\cite{nerf,Li_2023_ICCV,depth_nerf,barron2023zipnerf,barron2022mipnerf360,li2024nerfxl,chugunov2024light,NEURIPS2022_fe989bb0} still struggle with real-time rendering, as they require multiple queries to render a single pixel, which limits their practical use. Recently, 3D Gaussian Splatting (3DGS)~\cite{3dgs} has emerged as a more efficient alternative, enabling real-time NVS with performance comparable to top NeRF methods like Mip-NeRF 360\cite{barron2022mipnerf360}. The efficiency of 3DGS comes from its explicit scene representation, using learnable anisotropic 3D Gaussians and differentiable splatting with tile-based rasterization.

Building on this, several works have been introduced to improve the rendering quality, robustness, and generalization of 3DGS, as well as to explore its applications in specific areas~\cite{charatan23pixelsplat,Huang2DGS2024,chen2024mvsplat,guedon2023sugar,liu2025mvsgaussian,CVPR24_splatter_image,CVPR24_GS_IR,X_Ray_GS_ECCV,kheradmand20243d,Fu_2024_CVPR}. For example, 2DGS~\cite{Huang2DGS2024} simplifies the 3D volume into 2D oriented planar Gaussian disks to capture more accurate surface normals. MVSplat~\cite{chen2024mvsplat} and MVSGaussian~\cite{liu2025mvsgaussian} focus on achieving generalizable, feed-forward NVS. Additionally, X-Gaussian~\cite{X_Ray_GS_ECCV} applies 3DGS to X-ray NVS, demonstrating its effectiveness in sparse-view CT reconstruction.

\subsection{Curve Adjustment in Image Processing}
Curve-based adjustment modifies an image's tonal range or brightness by altering pixel values with a tone curve, which is commonly applied in commercial image software like Adobe Lightroom$^\circledR$. In earlier time, predefined curves (e.g., power curves, S-curves) were often used, with experts manually adjusting them to enhance detail and align with human visual perception~\cite{Siggraph2002_curve,curve_cvpr2010,Curve_old,CVPR11_curve}.

\begin{figure}
    \centering
    \includegraphics[width=0.98\linewidth]{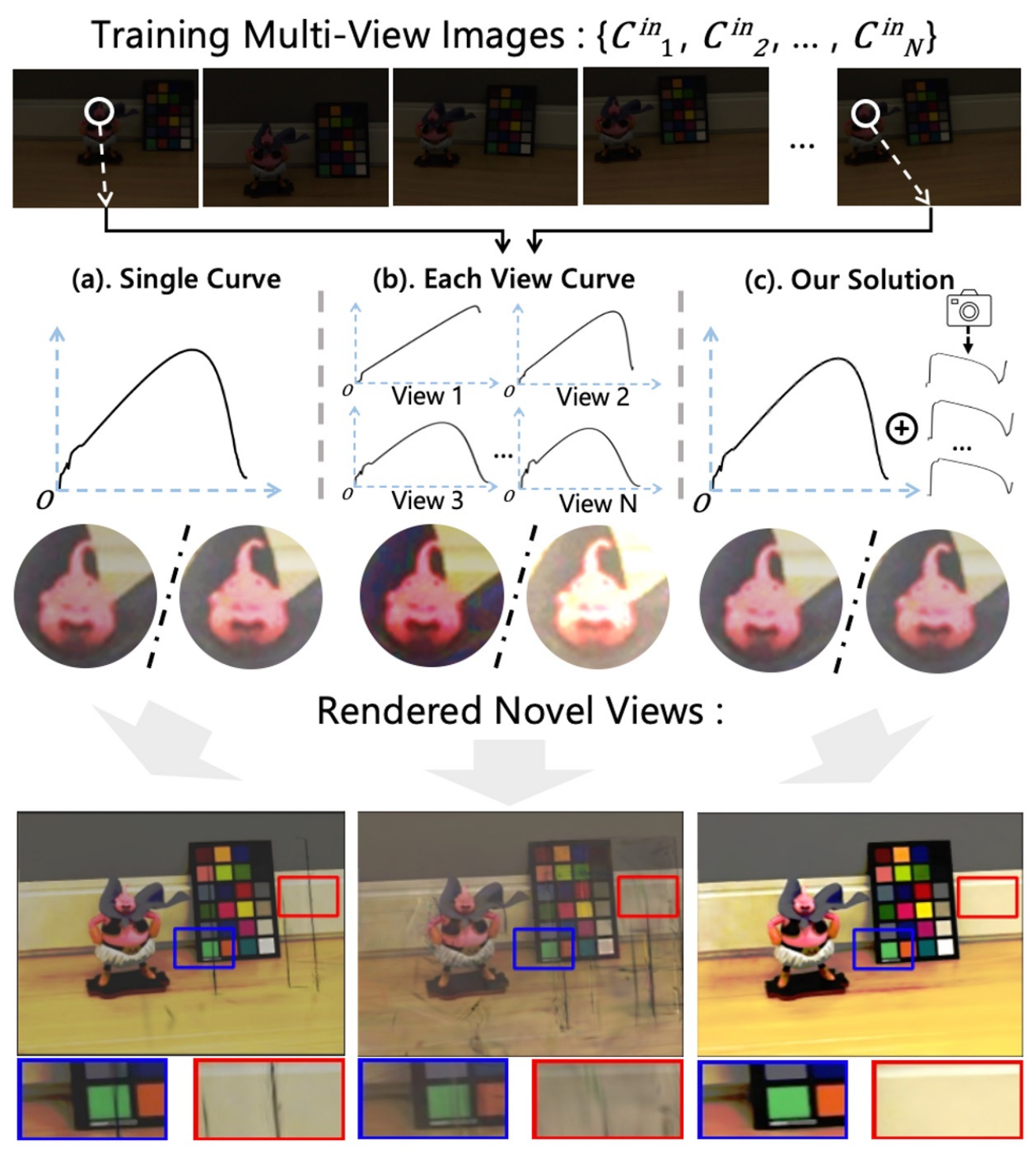}
    \vspace{-2mm}
    \caption{Ablation analysis of different curve settings in LOM dataset~\cite{cui_aleth_nerf} \textit{\textbf{``buu"}} scene: (a). All views low-light images share a single curve, (b). each view low-light image adopts an individual trainable curve and (c). our Luminance-GS solutions.}
    \label{fig:curve}
    \vspace{-4mm}
\end{figure}

In the deep learning era, data-driven solutions have become popular, using curves learned from datasets to perform global or local adjustments on images~\cite{Zero-DCE,SACC_2022_ACMMM,jiang2023meflut,moran2020curl,Curve_prediction_ECCV,curve_TOG,yang2023difflle,eccv24_colorcurves,lee2024cliptone,cuibmvc2024}. For example, CURL~\cite{moran2020curl} learns curves across multiple color spaces (RGB, HSV, CIELab) for targeted adjustments, and Zero-DCE~\cite{Zero-DCE} estimates pixel-wise, high-order curves, leveraging unsupervised loss functions for low-light enhancement. Recently, NamedCurves~\cite{eccv24_colorcurves} decomposes images based on the color naming system and applies Bézier curves for each color category. Unlike these 2D methods, our approach applies view-dependent curve adjustments across multiple views. We aim to correct brightness and contrast while ensuring 3D consistency, facilitating stable 3DGS training.

\begin{figure*}
    \centering
    \includegraphics[width=1.00\linewidth]{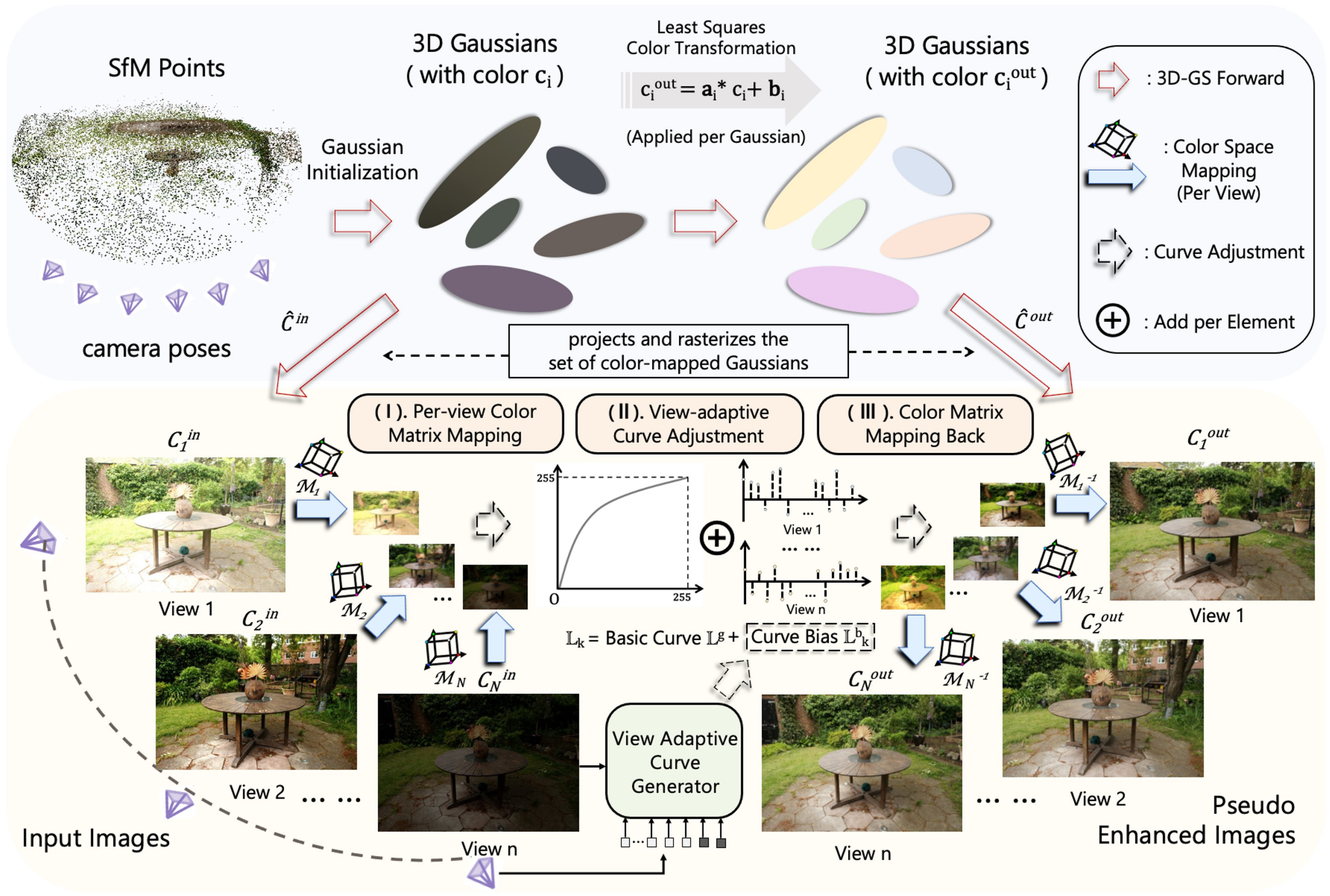}
    \vspace{-6mm}
    \caption{\textbf{Overview of Luminance-GS pipeline}. \textbf{\textit{Up}}: Our method jointly optimize 3D Gaussians with two set of color attributes $c_i$ and $c_i^{out}$ to render out input images $C^{in}$ and pseudo enhanced images $C^{out}$. \textbf{\textit{Down}}: To translate $C^{in}$ in to view-aligned enhanced $C^{out}$, we design 3 steps: (\uppercase\expandafter{\romannumeral1}). per-view color matrix mapping, (\uppercase\expandafter{\romannumeral2}). view-adptive curve adjustment and (\uppercase\expandafter{\romannumeral3}). color matrix mapping back.} 
    \label{fig:overview}
    \vspace{-4mm}
\end{figure*}

\subsection{NVS in Challenging Lighting Conditions}


While NeRF and 3DGS-based methods have achieved significant success in novel view synthesis (NVS) under normal lighting, a parallel line of research has emerged to enhance NVS performance under diverse, challenging real-world lighting conditions. Methods like NeRF-W~\cite{martinbrualla2020nerfw} and follow-up works~\cite{HA-NeRF,yang2023CRNeRF,GS-W_ECCV2024,Dahmani2024SWAGSI,kulhanek2024wildgaussians} address NeRF rendering under inconsistent lighting and the presence of occluding objects. For low-light conditions, RAW-NeRF~\cite{raw_nerf} and its extensions~\cite{zhang2024darkgs,jin2024le3d,li2024chaosclarity3dgsdark,singh24_hdrsplat} improve details by training with high dynamic range (HDR) camera RAW data. However, these methods are constrained by challenges in RAW data capture and storage, as well as longer training times.



Similar to our work, several studies focus on addressing NVS in challenging lighting conditions using sRGB inputs~\cite{LLNeRF,cui_aleth_nerf,xu2024leveraging,zou2024enhancing_aaai,ye2024gaussian_in_dark,TOG_wang2024bilateral,qu2024lushnerf}, like
Aleth-NeRF~\cite{cui_aleth_nerf} introduces the concept of ``concealing field" to control lightness, BilaRF~\cite{TOG_wang2024bilateral} optimizes 3D bilateral grids to simulate camera pipeline effects for each view, and Thermal-NeRF~\cite{xu2024leveraging} fuses low-light sRGB images with thermal images to achieve normal-light reconstruction results. Unlike previous methods, our Luminance-GS focuses more on using curve adjustment to generate and optimize pseudo-enhanced images. Moreover, our method can adapt to varying different challenging lighting conditions without modifying any training strategies or hyper-parameters.


%





\section{Proposed Method}
\label{sec:methods}


\subsection{Preliminary: 3D Gaussian Splatting}

3D Gaussian Splatting (3DGS)~\cite{3dgs} is an explicit point-based 3D representation with a set of Gaussians $\left\{ G_1, ..., G_M \right\}$. Each Gaussian $G_{i}$ is 
characterized by center position $\mu_i$, covariance matrix $\Sigma_i$, spherical harmonics coefficients $c_i$ (which represent color) and opacity $o_i$.
During rendering, each Gaussian will be projected and accumulated on the 2D image plane with camera parameters~\cite{Surface_splatting}:
\begin{equation}
    \hat{C}(x) = \sum_{i=1}^N c_i o_i' \prod_{j=1}^{i-1} (1 - o_j'),
    \label{Eq:rendering}
\end{equation}
where $N$ is the set of Gaussians (sorted along the depth after projection) that affect the pixel $x$, The $o_i'$ is the opacity after multiplying with projected $\Sigma_i$.
Then 3D Gaussians will be optimized through a mixed loss function $\mathcal{L}_{\rm 3DGS}$ between the predicted image $\hat{C}$ and ground truth $C$:
\begin{equation}
    \mathcal{L}_{\rm 3DGS}(\hat{C}, C) = \lambda \mathcal{L}_{\rm DSSIM}(\hat{C}, C) + (1 - \lambda) \mathcal{L}_1(\hat{C}, C),
    \label{Eq:3d_gs_loss}
\end{equation}
where $\mathcal{L}_{\rm DSSIM}$ denotes DSSIM loss and $\mathcal{L}_1$ denotes L1 loss, $\lambda$ is a balancing weight. We refer more details such as adaptive Gaussian densification to original 3DGS paper~\cite{3dgs}.

\subsection{Luminance-GS Pipeline}

Fig.~\ref{fig:overview} provides an overview of \textbf{Luminance-GS}. Beginning with SfM points, we model a set of 3D Gaussians, $\left\{ G_1, \dots, G_M \right\}$, with color attributes $c_i$ to render multi-view images $\left\{ C^{in}_1, C^{in}_2, \dots, C^{in}_N \right\}$, captured under challenging lighting conditions (e.g., low-light, varying exposure).


Within original 3DGS framework, we additional learn a set of color adjustment parameters $\mathbf{a}_i$ and $\mathbf{b}_i$, jointly
optimized with other 3DGS parameters.  $\mathbf{a}_i$ and $\mathbf{b}_i$ transform color
$c_i$ into $c_i^{out}$ using a least-squares formula~\cite{kulhanek2024wildgaussians}:
\begin{equation}
    c_i^{out} = \mathbf{a}_i \cdot c_i + \mathbf{b}_i,
\end{equation}
3D Gaussians $G_{i\in(1, M)}$ are simultaneously applied alongside the transformed colors $c_i^{out}$
 to render pseudo-enhanced images  $C^{out}$, which are translated from the input images $C^{in}$. During training time, Luminance-GS jointly predicts both $C^{in}$ and $C^{out}$, leading to an extension of Eq.\ref{Eq:rendering} as:
\begin{equation}
\begin{aligned}
    \hat{C}^{in}(x) &= \sum_{i=1}^N c_i o_i' \prod_{j=1}^{i-1} (1 - o_j') \\
    \hat{C}^{out}(x) &= \sum_{i=1}^N c_i^{out} o_i' \prod_{j=1}^{i-1} (1 - o_j').
\label{Eq:rendering_ours}
\end{aligned}
\end{equation}
During inference time, 3D Gaussians would only applied alongside $c_i^{out}$ to render out novel view $\hat{C}^{out}(x)$. In following section, we would explain how to convert the input images $\left\{ C^{in}_1, C^{in}_2, \ldots, C^{in}_N \right\}$ into view-aligned, pseudo-enhanced images $\left\{ C^{out}_1, C^{out}_2, \ldots, C^{out}_N \right\}$.

\begin{figure*}
    \centering
    \includegraphics[width=0.95\linewidth]{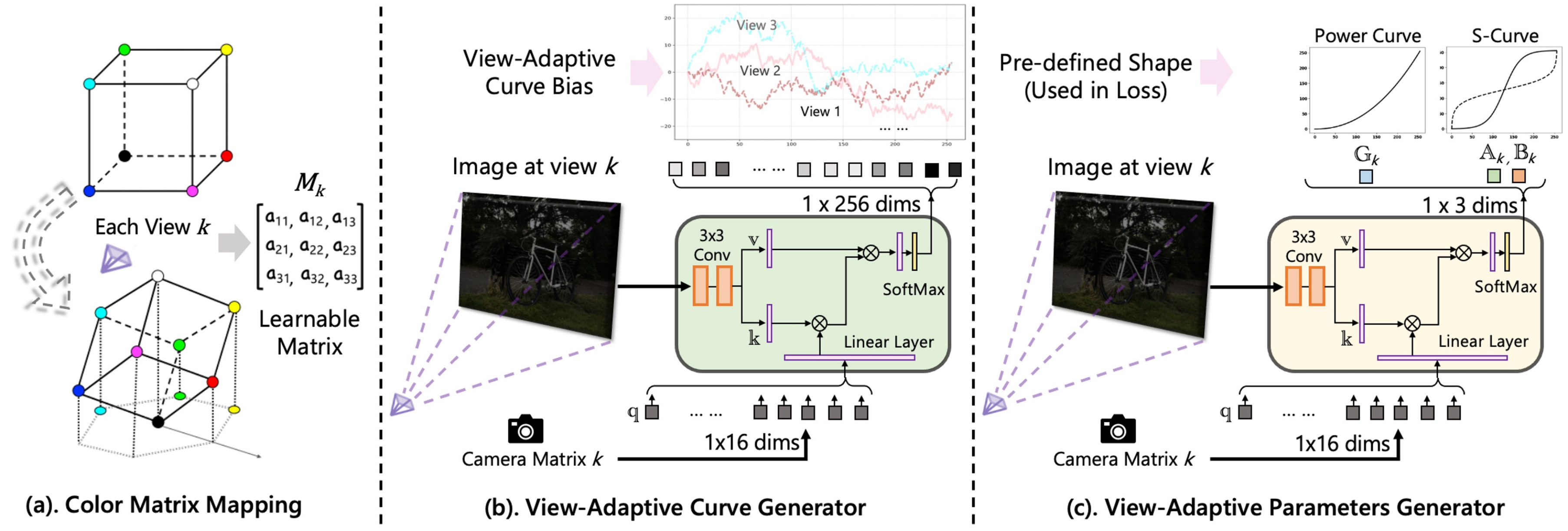}
    \vspace{-2mm}
    \caption{(a). Per-view Color Matrix Mapping with learnable matrix $\mathcal{M}_k$. (b). Structure of view-adaptive curve generator, given input image $C^{in}_k$ and corresponding camera pose to predict curve bias ${\mathbb L}_k^b$. (c). Structure of view-adaptive parameters generator.}
    \label{fig:detail_block}
    \vspace{-3mm}
\end{figure*}

\subsubsection{Problem Setup}

Given multi-view images $C^{in}_{k \in (1, N)}$
  captured under challenging lighting conditions (e.g., low-light, overexposure, or varying exposure), our primary goals for effective 3DGS training—which requires uniform lighting for optimal performance—are: (1) restoring lighting to approximate normal conditions, and (2) ensuring consistent illumination across views to preserve multi-view consistency in 3D. We adopt a straightforward curve adjustment process to achieve these two goals, the basic operation maps input pixel values to corresponding output values, enabling brightness and contrast adjustments~\cite{CVPR11_curve,Curve_old,eccv24_colorcurves}. By applying tone curves to map the input images $C^{in}_{k \in (1, N)}$ to a set of uniformly normal-light images $C^{out}_{k \in (1, N)}$, we aim to ensure color and luminance consistency across views $k \in (1, N)$, thereby facilitating 3DGS novel view synthesis.



In our implementation, we initially applied a single tone curve to all input views $C^{in}_{k \in (1, N)}$, however, as illustrated in Fig.~\ref{fig:curve}(a), even under relatively consistent low-light condition, single curve approach introduced view inconsistencies and rendering artifacts. We then experimented with assigning a unique curve to each view, but this only worsened the problem (see Fig.~\ref{fig:curve}(b)), the individual curves quickly became unstable and overfitted during training, resulting in pronounced inter-view discrepancies and significantly degraded rendering quality. To address this, we developed two solutions: Per-view Color Matrix Mapping and View-adaptive Curve Adjustment. With additional loss constraints, our Luminance-GS model effectively resolves these issues, achieving high-quality rendering results (see Fig.~\ref{fig:curve}(c)). The detailed explanations are as follows.

\subsubsection{Per-view Color Matrix Mapping}


In multi-view scenes, differences in lighting and camera exposure settings often result in color variations among images~\cite{GS-W_ECCV2024,kulhanek2024wildgaussians}. To address this, we first learn a unique matrix $\mathcal{M}_k$
  for each view $k$. Before curve adjustment, each input image is projected into its suitable view-dependent coordinate system via $\mathcal{M}_k$~\cite{cuibmvc2024}. A demonstration of color matrix mapping can be found in Fig.~\ref{fig:detail_block}(a).

As indicated by the blue arrow in Fig.~\ref{fig:overview}, given input image $C^{in}_{k \in (1, N)}(r,g,b)$ in the RGB color space $\mathbb{I}(R, G, B)$, for each view $k$, we optimize a per-view  3 $\times$ 3 invertible matrix $\mathcal{M}_k$.  Initially set as an identity matrix, $\mathcal{M}_k$ is jointly optimized with other 3DGS parameters. We define the color matrix mapping process as $F$, shown as follow equation:
\begin{equation}
\begin{aligned}
  F(C^{in}_k)  &= C^{in}_k \cdot \mathcal{M}_k
   \\
  &= \left[C^{in}_k(r), C^{in}_k(g), C^{in}_k(b) \right] \cdot \begin{bmatrix}
   a_{11} & a_{12} & a_{13}  \\
   a_{21} & a_{22} & a_{23} \\
   a_{31} & a_{32} & a_{33} 
  \end{bmatrix} \\
  &= \left[C^{in}_k(r'), C^{in}_k(g'), C^{in}_k(b') \right], 
\end{aligned}
\label{eq:matrix_trans}
\end{equation}
$F(C^{in}_k)$ is the output in view $k$, where RGB channels of input image $C^{in}_k(r), C^{in}_k(g), C^{in}_k(b)$ are transformed by matrix $\mathcal{M}_k$. Each element $\mathcal{M}_k (a_{ij})$ is a learnable parameter that enables color matrix mapping to adjust view characteristics. The transformed values $C^{in}_k(r'), C^{in}_k(g'), C^{in}_k(b')$, represent the image in the projected coordinate system, ready for subsequent view-adaptive curve adjustments.

\subsubsection{View-adaptive Curve Adjustment}

For curve design, we first learn a global curve ${\mathbb L}^g$ that is shared across all views $k \in (1, N)$. Then, for each view $k$, we learn a curve bias ${\mathbb L}_k^b$, which is added to the global curve to obtain the final tone curve ${\mathbb L}_k$: ${\mathbb L}_k = {\mathbb L}^g + {\mathbb L}_k^b$. Here the global curve ${\mathbb L}^g$ helps keep brightness and color tone consistent across all views, which works well in evenly lit environments (see Table.\ref{tab:ablation}). Additionally, the view-specific curve bias ${\mathbb L}_k^b$ 
  allows each view to fine-tune its brightness, providing flexibility for individual adjustments. 


In our implementation, global curve 
${\mathbb L}^g$  is set as a one-dimensional parameter with a length of 256 (ranging from 0 to 255). For curve bias ${\mathbb L}_k^b$, to avoid overfitting, we incorporate an attention block as a view-adaptive curve generator, inspired by~\cite{DETR,Cui_2022_BMVC}. The detailed structure is shown in Fig.~\ref{fig:detail_block}(b). For each view $k$, image $C^{in}_k$ is encoded by two down-scaling convolution layers and two linear layers to produce attention’s key and value. Then camera matrix at view $k$ would set as query to calculate cross attention~\cite{dosovitskiy2021an} with key and value, followed by a feed-forward network (FFN) that projects the output to a 1$\times$256 dimension ${\mathbb L}_k^b$. 

Afterward, ${\mathbb L}_k^b$ is added to ${\mathbb L}^g$ to produce ${\mathbb L}_k$, which is then applied to map the image 
$F(C^{in}_k)$ to the output value ${\mathbb L}_k(F(C^{in}_k))$ Finally, the result is multiplied by  the inverse of $\mathcal{M}_k$ and mapped back to the RGB color space, generating the pseudo-enhanced image $C^{out}_k$, as follow:
\begin{equation}
    C^{out}_k = {\mathbb L}_k(C^{in}_k \cdot \mathcal{M}_k) \cdot \mathcal{M}_k^{-1}.
\end{equation}
In next section, we would introduce how to set effective loss functions to control curve ${\mathbb L}_k$ and our model training.

\subsection{Optimization Solution}


\subsubsection{Image-level Loss Constraints}
We begin by introducing the image-level loss functions, as shown in Eq.\ref{Eq:rendering_ours}, Luminance-GS jointly predicts the input multi-view images $C^{in}_{k \in (1, N)}$ and the pseudo-enhanced images $C^{out}_{k \in (1, N)}$ consequently, the regression loss $\mathcal{L}_{\rm reg}$ for Luminance-GS is updated from Eq.\ref{Eq:3d_gs_loss} as follows:
\begin{equation}
    \mathcal{L}_{\rm reg} = \mathcal{L}_{\rm 3DGS}(\hat{C}^{in}, C^{in}) + \mathcal{L}_{\rm 3DGS}(\hat{C}^{out}, C^{out}).
\end{equation}
Secondly, inspired by Zero-DCE~\cite{Zero-DCE}, we introduce loss function $\mathcal{L}_{\rm spa}$, to preserve structural similarity between the predicted enhanced image $\hat{C}^{out}$ and the input image $C^{in}$. $\mathcal{L}_{\rm spa}$ will compute the difference between the neighboring pixel distances of $\hat{C}^{out}$ and $C^{in}$, equation as follow:
\begin{equation}
\begin{aligned}
    \mathcal{L}_{\rm spa} = & \frac{1}{K}\sum\limits_{x=1}^K\sum\limits_{y\in\Omega(x)}(|(\hat{C}^{out}(x)-\hat{C}^{out}(y))|\\
    & - \frac{0.5}{\overline{C^{in}(x)}} |(C^{in}(x)-C^{in}(y))|)^2.
\end{aligned}
\end{equation}
Here, $x$ denotes the pixel location, and $y \in \Omega(x)$ represents neighboring pixels of $x$. Unlike~\cite{Zero-DCE,cui_aleth_nerf}, which set difference parameter to be fixed, we treat it as an adaptive parameter $\frac{0.5}{\overline{C^{in}(x)}}$, where $\overline{C^{in}(x)}$ denotes the mean pixel value of the current view image $C^{in}(x)$, which better ensures the handling of input images with varying lighting conditions.



\subsubsection{Curve-level Loss Constraints}
Beyond image-level losses, we introduce curve-level loss constraints on $\mathbb{L}$ to control both its values and shape, which is essential for generating pseudo-enhanced images $C^{out}$ and for overall Luminance-GS training stability.


To prevent excessive divergence of the curve 
$\mathbb{L}$, we incorporate a shape regularization term that guides the curve to approximate specific patterns: the Power Curve ${\mathbb{L}}_{po}$, and the S-Curve ${\mathbb{L}}_s$, equation as follows. This regularization term serves as a prior, effectively restricting the curve's shape and ensuring that the generated images remain smooth (see supplementary). Equation as follows:
\begin{equation}
\begin{aligned}
 \mathbb{L}_{po}: y &= (x + \epsilon)^{\mathbb G_k}, \epsilon =1e^{-4}  \qquad \qquad \qquad 0 \leq x \leq 1 \\
 \mathbb{L}_{s}: y &= \begin{cases}
\mathbb A_k - \mathbb A_k \cdot \left( 1 - \frac{x}{\mathbb A_k} \right)^{\mathbb B_k}, & \text{if } 0 \leq x \leq \mathbb A_k\\
\mathbb A_k + (1 - \mathbb A_k) \cdot \left( \frac{x - \mathbb A_k}{1 - \mathbb A_k} \right)^{\mathbb B_k}, & \text{if } 1 \geq x > \mathbb A_k
\end{cases}
\end{aligned}
\end{equation}
where $\left\{\mathbb G_k, \mathbb A_k, \mathbb B_k \right\}$ are the view-adaptive learnable parameters, same as the view-adaptive curve generator, we additionally design another view-adaptive parameter generator to learn and predict $\left\{\mathbb G_k, \mathbb A_k, \mathbb B_k \right\}$ (see Fig.~\ref{fig:detail_block}(c)). Additionally, to control curve value, we use cumulative distribution function (CDF) $\mathbb{L}_{\rm cdf}$ of the histogram-equalized (HE) input image $C^{in}(x)$ as the initial target values, the curve loss $\mathcal{L}_{\rm curve}$ is set to learn both $\mathbb{L}_{\rm cdf}$ and shaped parameters:  
\begin{equation}
    \mathcal{L}_{\rm curve} = \omega ||\mathbb{L} -  \mathbb{L}_{\rm cdf}||^2  + 0.5 ||\mathbb{L} - (\mathbb{L}_{po}\cdot\mathbb{L}_{s}) ||^2 
\end{equation}
where $\omega$ is a weight parameter, set to 1.0 for the first 3,000 iterations and set to 0.1 after 3,000 iterations,  please refer to our supplementary for more details. Additionally, a total variation (TV) loss is incorporated between adjacent values in $\mathcal{L}$ to enforce smoothness, defined as follows:
\begin{equation}
    \mathcal{L}_{\rm tv} = (1/255) \cdot \sum_{i \in (0,254)} |\mathcal{L}(i+1) - \mathcal{L}(i)|^2
\end{equation}
Finally, the total loss of Luminance-GS is defined as:
\begin{equation}
    \mathcal{L}_{\rm total} = \mathcal{L}_{\rm reg} + \mathcal{L}_{\rm spa} + \mathcal{L}_{\rm tv} + 10 \cdot \mathcal{L}_{\rm curve}.
\end{equation}

\section{Experiments}
\label{sec:exps}



\begin{table}[]
\caption{Mean results of LOM dataset~\cite{cui_aleth_nerf} low-light subset (PSNR $\uparrow$, SSIM $\uparrow$, LPIPS $\downarrow$). Best results are bolded, second-best underlined, per-scene results please refer to supplementary.}
\vspace{-2mm}
\label{tab:LOM_low}
\LARGE
\centering
\renewcommand\arraystretch{1.8}
\begin{adjustbox}{max width = 1\linewidth}
\begin{tabular}{l|c|l|c}
\toprule
\toprule
Method       & PSNR/ SSIM/ LPIPS   & Method       & PSNR/ SSIM/ LPIPS   \\ 
\midrule
\midrule
GS~\cite{3dgs}  & 6.88/ 0.157/ 0.662  & GS + Z-DCE~\cite{Zero-DCE} & 13.64/ 0.672/ 0.408 \\ \hline
Z-DCE~\cite{Zero-DCE} + GS & 13.45/ 0.702/ 0.349 & GS + SCI~\cite{cvpr22_sci}   & 15.22/ 0.748/ 0.430 \\ \hline
SCI~\cite{cvpr22_sci} + GS   & 11.73/ 0.692/ 0.407 & GS + NeRCo~\cite{ICCV2023_NeRCo} & 17.21/ 0.712/ 0.421 \\ \hline
NeRCo~\cite{ICCV2023_NeRCo} + GS & 17.59/ 0.727/ \underline{0.345} & LLVE~\cite{LLVE_2021_CVPR} + GS  & 16.43/ 0.728/ 0.399 \\ \hline
SGZ~\cite{SGZ_wacv2022} + GS   & 14.14/ 0.706/ 0.353 & AME-NeRF~\cite{zou2024enhancing_aaai}     & 17.65/ 0.729/ 0.405 \\ \hline
Aleth-NeRF~\cite{cui_aleth_nerf}   & \textbf{19.87}/ \underline{0.754}/ 0.417 & \textbf{Luminance-GS} & \underline{18.34}/ \textbf{0.799}/ \textbf{0.294} \\ 
\bottomrule
\bottomrule

\end{tabular}
\end{adjustbox}
\vspace{-4mm}
\end{table}

\begin{figure*}
    \centering
    \includegraphics[width=0.95\linewidth]{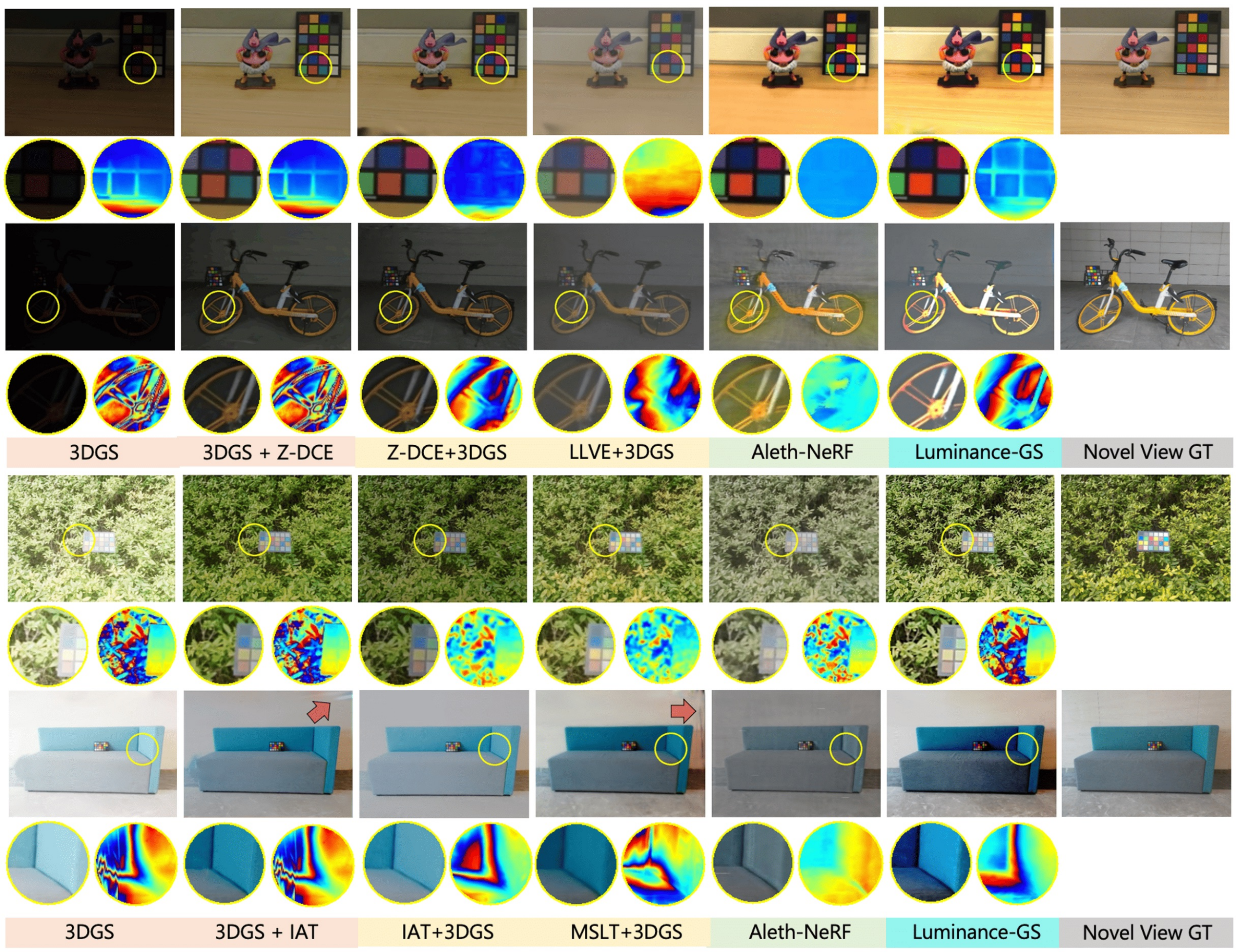}
    \vspace{-1.5mm} 
    \caption{Novel view synthesis results on LOM dataset~\cite{cui_aleth_nerf} low-light \textbf{\textit{``buu"}}, \textbf{\textit{``bike"}} scenes and overexposure \textbf{\textit{``shrub"}}, \textbf{\textit{``sofa"}} scenes.}
    \label{fig:LOM_exp}
    \vspace{-4mm} 
\end{figure*}

Our code is based on open-source toolbox GS-Splat~\cite{tool_box}. We evaluate our model under 3 different lighting conditions: \textit{(a)} low-light, \textit{(b)} overexposure, and \textit{(c)} varying exposure. We present the dataset information and comparative methods, followed by experimental results and ablation details.

\textbf{Dataset:} For \textit{(a)} low-light condition, we adopt low-light subset of the LOM dataset~\cite{cui_aleth_nerf}, which includes five scenes: “\textbf{\textit{buu}},” “\textbf{\textit{chair}},” “\textbf{\textit{sofa}},” “\textbf{\textit{bike}},” and “\textbf{\textit{shrub}}.” Each scene provides low-light base views for training and normal-light novel views for evaluation. For \textit{(b)} overexposure condition, we use the overexposure subset of the LOM dataset, which includes the same scenes but provides overexposed views for training and normal-light views for evaluation.
For \textit{(c)} varying exposure condition, we adopt unbounded Mip-NeRF 360 dataset~\cite{barron2022mipnerf360}, we synthesize each training views with different exposure conditions~\cite{Afifi_2021_CVPR}, meanwhile we also apply a slight gamma adjustment to each view to increase the difficulty (see Fig.~\ref{fig:motivation} and Fig.~\ref{fig:overview} for examples). The unbounded $360^\circ$ scene and varying exposure conditions would make NVS task more challenging.

\textbf{Comparison Methods:} 
For \textit{(a)} low-light condition, we first compare with original 3DGS~\cite{3dgs}. Then, we compare with combination of SOTA 2D image $\&$ video enhancement methods (Z-DCE~\cite{Zero-DCE}, SCI~\cite{cvpr22_sci}, NeRCo~\cite{ICCV2023_NeRCo}, SGZ~\cite{SGZ_wacv2022}, LLVE~\cite{LLVE_2021_CVPR}) with 3DGS, includes using enhancement methods to pre-process images for training (`` '' + 3DGS in Table \ref{tab:LOM_low}) and applying enhancement methods to post-processing images during rendering (3DGS + `` '' in Table \ref{tab:LOM_low}). Finally, we compare with two NeRF-based enhancement methods: AME-NeRF~\cite{zou2024enhancing_aaai} and Aleth-NeRF~\cite{cui_aleth_nerf}. 


For \textit{(b)} overexposure condition, we compare with 3DGS~\cite{3dgs}, combination of 3DGS and 2D exposure correction methods (MSEC~\cite{Afifi_2021_CVPR}, IAT~\cite{Cui_2022_BMVC}, MSLT~\cite{zhou2024mslt}, Adobe Lightroom$^\circledR$ manually adjustment) and 3DGS, as well as NeRF-based exposure correction method Aleth-NeRF.


For \textit{(c)} varying exposure conditions, we compare with 3DGS~\cite{3dgs}, NeRF-based enhancement method Aleth-NeRF, NeRF-based in-the-wild method NeRF-W~\cite{martinbrualla2020nerfw}, and its recent 3DGS follow-up, GS-W~\cite{GS-W_ECCV2024}. Due to page limitation, we only show 3 scenes results “\textbf{\textit{bicycle}}”, “\textbf{\textit{garden}}” and “\textbf{\textit{counter}} here, full results please refer to supplementary.


\subsection{Experimental Results}

The \textit{(a)} low-light experimental results is shown in Table.~\ref{tab:LOM_low},  we show the mean results of all 5 scenes~\footnote{Per scene results is shown in our supplementary part.}, with comparison of various methods, our Luminance-GS achieved the best results in both SSIM and LPIPS, outperforming the previous SOTA method~\cite{cui_aleth_nerf} with an improvement of $\uparrow$ 0.44 in SSIM and $\downarrow$ 0.123 in LPIPS, although our PSNR is the second best, we achieve better detail recovery and more reliable depth maps (see Fig.~\ref{fig:LOM_exp}). The \textit{(b)} overexposure experimental results is shown in Table.~\ref{tab:LOM_oe}, our Luminance-GS achieved the best results in all metrics, showing significant improvement over other methods. Fig.~\ref{fig:motivation} showcases a comparison of efficiency. Some qualitative results can be found in Fig.~\ref{fig:LOM_exp}, compared to other approaches, our Luminance-GS effectively preserves 3D multi-view consistency meanwhile retains rich details and vivid color appearance.

\begin{table}[t]
\caption{Mean results of LOM dataset~\cite{cui_aleth_nerf} overexposure subset. Best results are bolded, second-best underlined.} 
\vspace{-2mm}
\label{tab:LOM_oe}
\LARGE
\centering
\renewcommand\arraystretch{1.8}
\begin{adjustbox}{max width = 1\linewidth}
\begin{tabular}{l|c|l|c}
\toprule
\toprule
Method     & PSNR/ SSIM/ LPIPS   & Method         & PSNR/ SSIM/ LPIPS   \\ 
\midrule
\midrule
GS~\cite{3dgs}         & 9.64/ 0.726/ 0.392  & GS + MSEC~\cite{Afifi_2021_CVPR}      & 19.56/ 0.805/ 0.382 \\ \hline
MSEC~\cite{Afifi_2021_CVPR} + GS  & 17.20/ 0.767/ 0.363 & GS + IAT~\cite{Cui_2022_BMVC}       & 20.23/ 0.821/ 0.347 \\ \hline
IAT~\cite{Cui_2022_BMVC} + GS   & 17.56/ 0.800/ 0.311 & GS + MSLT~\cite{zhou2024mslt}      & \underline{20.39}/ 0.815/ 0.345 \\ \hline
MSLT~\cite{zhou2024mslt} + GS  & 20.25/ \underline{0.824}/ \underline{0.262} & Lightroom + GS & 19.91/ 0.785/ 0.308 \\ \hline
Aleth-NeRF~\cite{cui_aleth_nerf} & 18.09/ 0.745/ 0.488 & \textbf{Luminance-GS}   & \textbf{20.71}/ \textbf{0.835}/ \textbf{0.222} \\ 
\bottomrule
\bottomrule
\end{tabular}
\end{adjustbox}
\vspace{-4.5mm}
\end{table}

\begin{figure*}[t]
    \centering
    \includegraphics[width=0.97\linewidth]{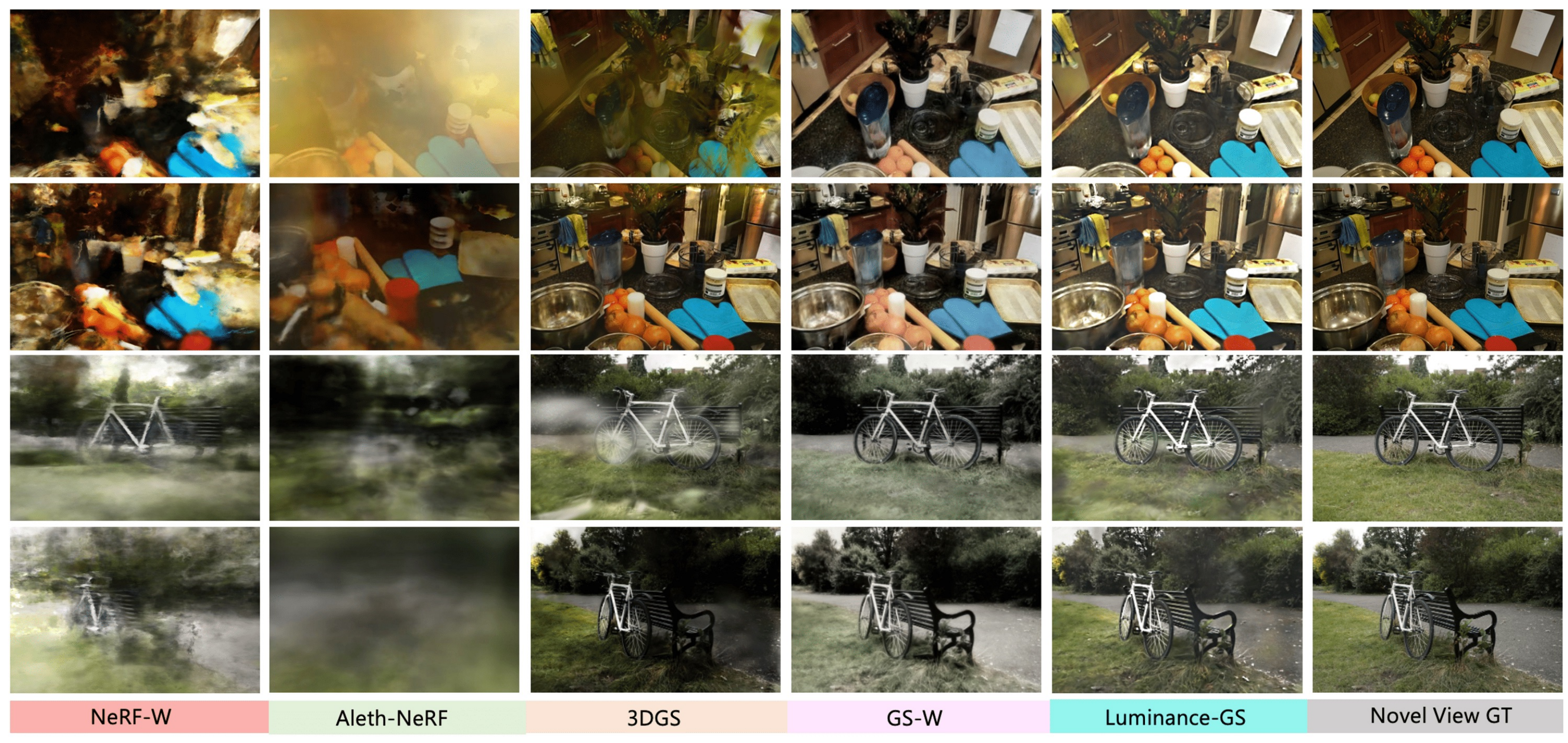}
    \vspace{-3mm}
    \caption{Novel view synthesis results on our synthesized varying exposure unbounded dataset (from Mip-NeRF 360 dataset~\cite{barron2021mipnerf}) “\textbf{\textit{counter}}” and “\textbf{\textit{bicycle}}” scenes, with comparison of NeRF-W~\cite{martinbrualla2020nerfw}, Aleth-NeRF~\cite{cui_aleth_nerf}, 3DGS~\cite{3dgs} and GS-W~\cite{GS-W_ECCV2024}.}
    \label{fig:MIP360_exp}
    \vspace{-4mm}
\end{figure*}

The \textit{(c)} varying exposure experimental results are shown in Table~\ref{tab:LOM_vary}. Our method achieves state-of-the-art (SOTA) image quality with rendering FPS on par with the original 3DGS~\cite{3dgs}, though with a slightly longer training time. The visualization results is shown in Fig.~\ref{fig:MIP360_exp}, we can find that Aleth-NeRF~\cite{cui_aleth_nerf} and 3DGS~\cite{3dgs} are unable to handle the variations in lighting and color, meanwhile NeRF-W~\cite{martinbrualla2020nerfw} faces challenges in unbounded scenes. Finally, compared to GS-W~\cite{GS-W_ECCV2024}, our method achieves better color recovery and results closer to the ground truth.

\begin{table}[t]
\caption{Comparison on varying exposure dataset, red color shows best result meanwhile yellow color shows second best result.} 
\vspace{-2mm}
\label{tab:LOM_vary}
\Huge
\centering
\renewcommand\arraystretch{1.9}
\begin{adjustbox}{max width = 0.95\linewidth}
\begin{tabular}{cl|ccccc}
\toprule[1.3pt]
\toprule[1.3pt]
\multicolumn{1}{c|}{scene}                    & \multicolumn{1}{c|}{metric} & 3DGS  \quad               & NeRF-W               & Aleth-NeRF           & \, GS-W \, \quad & \quad \textbf{Ours} \quad \quad                \\ \midrule \midrule
\multicolumn{1}{c|}{\multirow{3}{*}{\textit{\textbf{``bicycle"}}}} & PSNR $\uparrow$   & \cellcolor{red!20} 18.63 &  13.82 &  11.45 &  18.33 & \cellcolor{yellow!20}  18.52  \\
\multicolumn{1}{c|}{}                         & SSIM $\uparrow$    &   0.539 & 0.233 & 0.154 & \cellcolor{yellow!20} 0.618   & \cellcolor{red!20} 0.647 \\
\multicolumn{1}{c|}{}                         & LPIPS $\downarrow$                     & 0.409 &  0.631     &  0.772   &  \cellcolor{yellow!20}  0.377 & \cellcolor{red!20} 0.334   \\ \midrule \midrule
\multicolumn{1}{c|}{\multirow{3}{*}{\textit{\textbf{``garden"}}}}  & PSNR $\uparrow$   &   20.39  &  10.71  &  11.34  & \cellcolor{yellow!20} 20.49  & \cellcolor{red!20} 20.93 \\
\multicolumn{1}{c|}{}                         & SSIM $\uparrow$    &   0.755  &   0.222  &   0.245 &  \cellcolor{yellow!20} 0.765    & \cellcolor{red!20} 0.786 \\
\multicolumn{1}{c|}{}                         & LPIPS $\downarrow$    & \cellcolor{yellow!20}  0.205 &  0.687   &  0.758  & 0.222    & \cellcolor{red!20} 0.193  \\ \midrule \midrule
\multicolumn{1}{c|}{\multirow{3}{*}{\textit{\textbf{``counter"}}}} & PSNR $\uparrow$    &  14.01  & 10.81 &  9.87  & \cellcolor{yellow!20} 15.79  & \cellcolor{red!20} 16.29 \\
\multicolumn{1}{c|}{}   & SSIM $\uparrow$   &  0.491 & 0.411 &   0.249   & \cellcolor{yellow!20} 0.618 & \cellcolor{red!20} 0.631 \\
\multicolumn{1}{c|}{}   & LPIPS $\downarrow$     &  0.376 &  0.758    &  0.791  &  \cellcolor{yellow!20} 0.344    & \cellcolor{red!20} 0.313   \\ \midrule \midrule
\multicolumn{2}{c|}{GPU hrs $\downarrow$}                                                & \multicolumn{1}{c}{\cellcolor{red!20} $\sim$ 3min} & \multicolumn{1}{c}{$\sim$ 38hrs} & \multicolumn{1}{c}{$\sim$ 8hrs} & \multicolumn{1}{c}{$\sim$ 45min} & \multicolumn{1}{c}{\cellcolor{yellow!20} $\sim$ 18min} \\ \midrule \midrule
\multicolumn{2}{c|}{FPS $\uparrow$}    & \multicolumn{1}{l}{\cellcolor{red!20} $\sim$ 150}  & \multicolumn{1}{c}{0.138} & \multicolumn{1}{c}{0.875} & \multicolumn{1}{c}{\cellcolor{yellow!20}  $\sim$ 80} & \multicolumn{1}{c}{\cellcolor{red!20} $\sim$ 150} \\ \bottomrule[1.3pt] \bottomrule[1.3pt]
\end{tabular}
\end{adjustbox}
\vspace{-4mm}
\end{table}

\subsection{Ablation Analyze}
We conducted additional ablation studies to verify the effectiveness of each module in our curve design, we analyze the different part of curve $\mathbb L$, and the results is shown in Table.~\ref{tab:ablation}, we performed experiments on scenes under three different lighting conditions. From Table.~\ref{tab:ablation} we can find that the global curve ${\mathbb L}^g$ performs well in evenly lit environments (low-light, overexposure), meanwhile ${\mathbb L_k}^b$ plays a key role in varying exposure scenarios, our approach of combining ${\mathbb L}^g$ and ${\mathbb L_k}^b$ leads to improvements across all lighting scenarios. Meanwhile, the design of the projection matrix $\mathcal{M}_k$ has also proven effective in nearly all scenarios, demonstrating the effectiveness of per-view color matrix mapping. For more ablation analysis regarding the loss function, please refer to our supplementary material.

\begin{table}[t]
\caption{Ablation details of different parts in the curve design.} 
\vspace{-2mm}
\label{tab:ablation}
\Huge
\centering
\renewcommand\arraystretch{1.8}
\begin{adjustbox}{max width = 1\linewidth}
\begin{tabular}{l|ccc}
\toprule
\toprule
curve type  & \textit{\textbf{``buu''}} (low-light)     & \textit{\textbf{``buu''}} (overexposure)  & \textit{\textbf{``garden''}} (varying)    \\ \hline
${\mathbb L}^g$ only      & 17.98/ 0.796/ 0.211 & 19.00/ 0.789/ 0.343 & 14.51/ 0.573/ 0.579 \\
${\mathbb L}_k^b$ only    & 17.78/ 0.653/ 0.377 & 16.43/ 0.712/ 0.461 & 19.15/ 0.729/ 0.251 \\
${\mathbb L}^g$ + ${\mathbb L}_k^b$ & 18.05/ 0.859/ 0.198 & 19.37/ 0.804/ 0.344 & 20.26/ \textbf{0.790}/ 0.218 \\
${\mathbb L}^g$ + ${\mathbb L}_k^b$ + $\mathcal{M}_k$ & \textbf{18.09}/ \textbf{0.877}/ \textbf{0.193} & \textbf{19.67}/ \textbf{0.811}/ \textbf{0.311} & \textbf{20.93}/ 0.786/\textbf{0.193} \\ \bottomrule
\bottomrule
\end{tabular}
\end{adjustbox}
\vspace{-2mm}
\end{table}

\section{Conclusion}
\label{sec:conclusion}
In this paper, we propose \textbf{Luminance-GS}, which improves the robustness of 3D Gaussian Splatting under challenging lighting conditions, such as low-light, overexposure, and varying exposure. Compared with previous methods, Luminance-GS offers superior illumination generalization and speed advantages, while achieving state-of-the-art (SOTA) performance on both real-world and synthetic datasets. In the future, we aim to extend our method to achieve scene generalization, enabling it to handle both illumination and scene variability.



\section{Acknowledgment}

This research is partially supported by JST Moonshot R$\&$D Grant Number JPMJPS2011, CREST Grant Number JPMJCR2015 and Basic Research Grant (Super AI) of Institute for AI and Beyond of the University of Tokyo.

{
    \small
    \bibliographystyle{ieeenat_fullname}
    \bibliography{main}
}


\clearpage

\section{Detailed Experimental Results}

Due to page limitations, we could not fully present the per-scene experimental results in the main text. Instead, we provide the complete experimental results in the supplementary part: the per-scene results for the LOM dataset low-light scene are shown in Table.~\ref{tab:LOM_low_full}, it can be observed that our method achieves excellent performance in most scenarios. Additionally, we found that Aleth-NeRF~\cite{cui_aleth_nerf} is sensitive to hyper-parameter, for instance, if the enhance degree of Aleth-NeRF is set to a slightly lower value (e.g., reduced from 0.45 to 0.4), its PSNR value significantly decreases as well (see Table.~\ref{tab:LOM_low_full}).
The results for overexposure scenes in the LOM dataset are presented in Table.~\ref{tab:LOM_oe_full}, and the other 4 scenes results for our synthesized varying exposure dataset  are shown in Table.~\ref{tab:LOM_non_uni}. We can see that our Luminance-GS both achieves SOTA performance in PSNR, SSIM and LPIPS.

More visualization results are shown in Fig.~\ref{fig:supp_low}, Fig.~\ref{fig:supp_exposure} and Fig.~\ref{fig:supp_uneven}, we can found that sometimes image restoration modules easily lead to multi-view inconsistency, which ultimately causes floaters during rendering (see Fig.~\ref{fig:supp_exposure} ``MSEC~\cite{Afifi_2021_CVPR} + 3DGS'' for example). Meanwhile, our method achieves better detail reconstruction results compared to other approaches. We have zoomed in on random areas to enlarge the details and demonstrate the superior performance of our detail recovery.

\begin{figure}[t]
    \centering
    \includegraphics[width=1\linewidth]{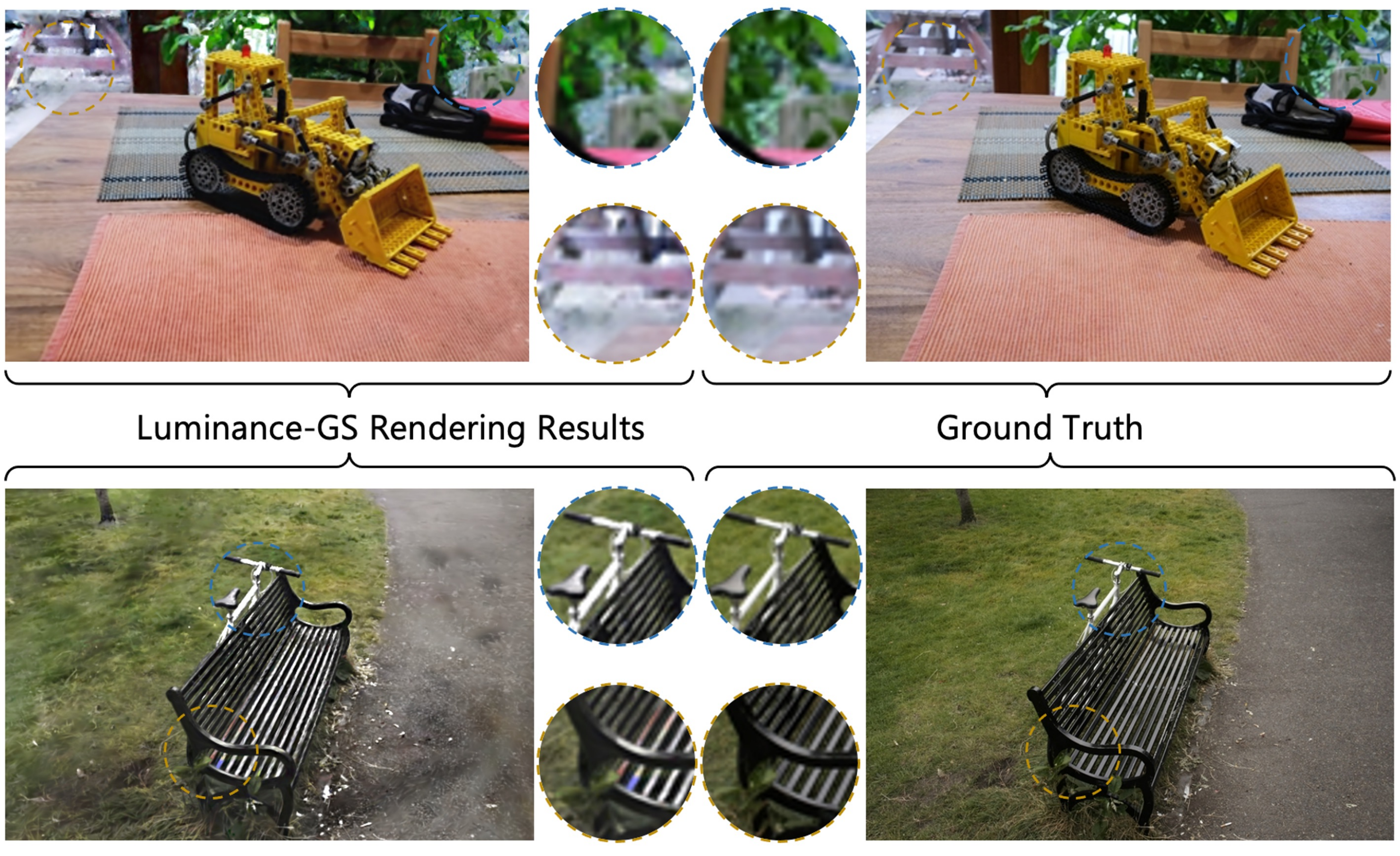}
    \caption{The limitation of our method, in some cases, Luminance-GS fails to correctly render colors, resulting in color discrepancies in certain areas and the occurrence of pixelation.}
    \label{fig:supp_limitation}
    \vspace{-3mm}
\end{figure}

\section{Ablation Analysis of Loss Functions}

We further assess the effectiveness of various loss functions in our Luminance-GS model. Figure~\ref{fig:supp_ablation} illustrates this through the \textbf{\textit{``sofa''}} scene, showcasing examples under both low-light and overexposure conditions.

From Fig.~\ref{fig:supp_ablation}, we can find that spatial loss $\mathcal{L}_{\rm spa}$ (Eq.8 in main text) plays a crucial role in maintaining multi-view consistency, after removing loss $\mathcal{L}_{\rm spa}$, the rendered scenes tend to exhibit large areas of floaters, which become particularly pronounced under low-light conditions. Meanwhile, in curve loss $\mathcal{L}_{\rm curve}$ (Eq.10 in main text), the cumulative distribution function (CDF) $\mathbb{L}_{\rm cdf}$ of the histogram-equalized (HE) $C^{in}(x)$  is essential for controlling the degree of illumination. Without $\mathbb{L}_{\rm cdf}$, efforts to enhance or attenuate illumination are often unsuccessful. Additionally, the pre-defined curve shape $\mathbb{L}_{po}\cdot\mathbb{L}_{s}$ (Eq.9 in main text) maintain the smoothness of the generated images, reducing the likelihood of large areas of pixels collapsing into a single value, ensuring the generated images more aligned with human visual perception. Ultimately, with the assistance of all the aforementioned losses, we can achieve satisfactory rendered novel views, as shown in the last column of Fig.~\ref{fig:supp_ablation}.

\section{Experimental Setup}

For training settings, we trained Luminance-GS on a single Nvidia Tesla V100 GPU using the Adam optimizer. The learning rates for the various parameters were set as follows:

\begin{figure*}[t]
    \centering
    \includegraphics[width=1\linewidth]{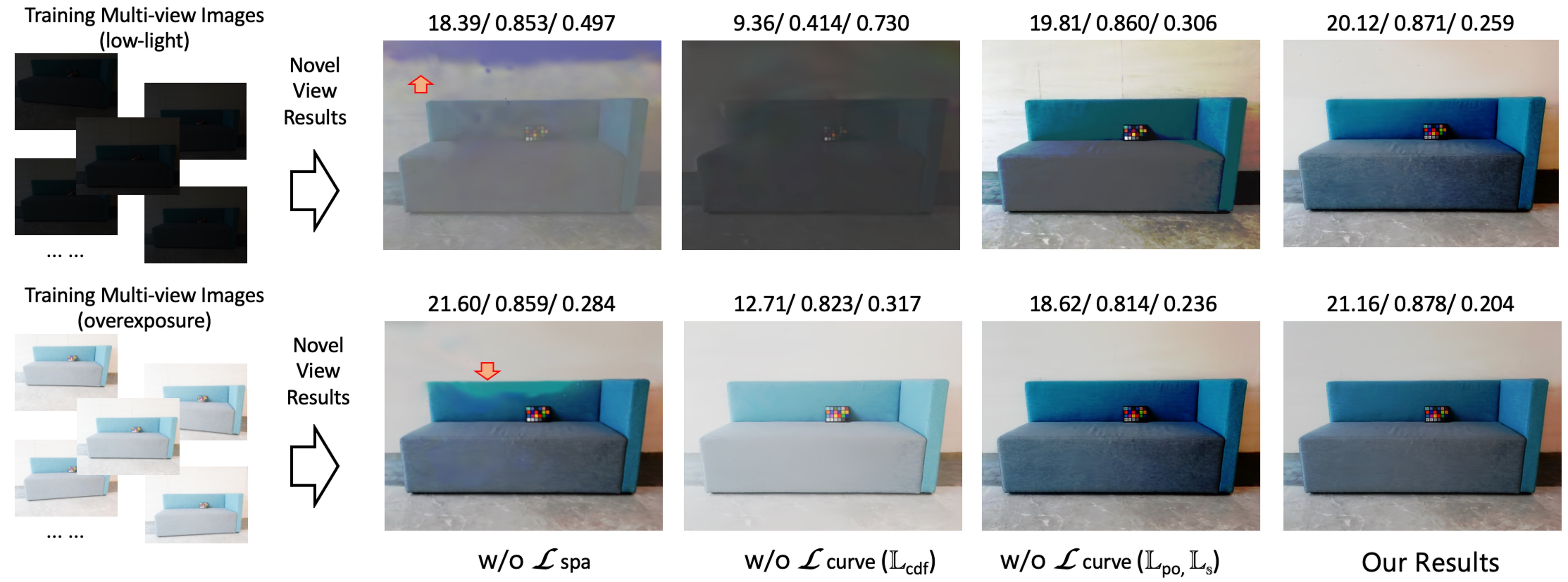}
    \caption{Ablation analysis  of different loss functions in Luminance-GS, we denote PSNR$\uparrow$/ SSIM$\uparrow$/ LPIPS$\downarrow$ value upon the figures.}
    \label{fig:supp_ablation}
    \vspace{-3mm}
\end{figure*}

\begin{itemize}
    \item For the basic 3DGS parameters $G_i$ = $\left\{ \mu_i, c_i, o_i, \Sigma_i \right\}$ , we adopted the default settings from GS-Splat~\cite{tool_box}.
    \item The learning rates for the color adjustment parameters $\mathbf{a}_i$ and $\mathbf{b}_i$ were set to $2.5 \times 10^{-3}$.
    \item The learning rate for the color space mapping matrix $\mathcal{M}_k$ was set to $2.5 \times 10^{-4}$, with a weight decay of $1 \times 10^{-5}$.
    \item The global curve $\mathbb{L}^g$ was optimized with a learning rate of $1 \times 10^{-3}$ and a weight decay of $1 \times 10^{-4}$.
    \item The learning rates for the two attention blocks (view-adaptive curve generator and view-adaptive parameters generator, as shown in Fig. 4 of the main text) were set to $1 \times 10^{-5}$, with a weight decay of $1 \times 10^{-5}$.
\end{itemize}
Training was conducted for a total of 10,000 iterations, with Gaussian refinement stopping at 8,000 iterations. For other settings, such as Gaussian reset steps~\cite{3dgs}, we adhered to the default configuration provided by GS-Splat~\cite{tool_box}.

\begin{table}[t]
\caption{Other 4 scenes results on varying exposure unbounded dataset, we show the PSNR $\uparrow$, SSIM $\uparrow$ and LPIPS $\downarrow$. Red indicates the best result, while blue indicates the second-best result.}
\vspace{-3mm}

\label{tab:LOM_non_uni}
\centering
\renewcommand\arraystretch{1.7}
\begin{adjustbox}{max width = 1\linewidth}
\huge
\begin{tabular}{c|c|c|c|c}
\toprule
\toprule
Methods    & \textbf{\textit{``bonsai"}} & \textbf{\textit{``kitchen"}} & \textbf{\textit{``room"}} & \textbf{\textit{``stump"}} \\ \midrule \midrule
3DGS       & 12.50/ 0.299/ 0.560 & \blue{22.26}/ \blue{0.832}/ 0.177 & 14.22/ 0.549/ \red{0.299} & \red{16.61}/ 0.462/ \blue{0.400} \\
NeRF-W     & 9.64/ 0.232/ 0.725 & 13.79/ 0.502/ 0.601 & 12.89/ 0.432/ 0.534 & 13.98/ 0.403/ 0.712 \\
Aleth-NeRF & 8.16/ 0.401/ 0.685 & 11.22/ 0.397/ 0.577 & 7.02/ 0.321/ 0.733 & 12.41/ 0.473/ 0.665 \\
GS-W       & \blue{15.46}/ \blue{0.541}/ \red{0.421} & 20.77/ 0.802/ \blue{0.167} & \red{17.19}/ \blue{0.558}/ 0.378 & 14.65/ \blue{0.503}/ 0.429 \\
Ours       & \red{15.61}/ \red{0.560}/ \blue{0.438} & \red{23.09}/ \red{0.834}/ \red{0.144} & \blue{16.77}/ \red{0.656}/ \blue{0.320} & \blue{15.63}/ \red{0.532}/ \red{0.388}  \\ \bottomrule \bottomrule
\end{tabular}
\end{adjustbox}
\end{table}

\section{Limitation and Future Discussion}

Some failure cases are shown in Fig.~\ref{fig:supp_limitation}. In certain scenes, Luminance-GS may lose fine details, such as the leaves of plants disappearing (see Fig.~\ref{fig:supp_limitation} above). Additionally, Luminance-GS sometimes renders incorrect colors and can exhibit pixelated artifacts, as seen in the chair's color in Fig.~\ref{fig:supp_limitation} below. This could be due to errors in the pseudo-labels generated by curve $\mathbb{L}$, and we hope that future research can optimize both the training strategy and the pseudo-label generation solution.

For future research directions, we think that it would be valuable to consider more scenarios of internal camera degradation, such as inconsistent white balance settings. Exploring how to enable scene generalization with Luminance-GS is also a promising direction. Additionally, we believe that extending Luminance-GS to real-world applications, such as autonomous driving and underground coal mining scenarios, would be highly meaningful.

\begin{table*}[t]
\caption{Per-scene experimental results (PSNR $\uparrow$, SSIM $\uparrow$, LPIPS $\downarrow$) on LOM dataset~\cite{cui_aleth_nerf} low-light subset, we compare with various enhancement methods~\cite{Zero-DCE,cvpr22_sci,ICCV2023_NeRCo,LLVE_2021_CVPR,SGZ_wacv2022} and NeRF-based methods~\cite{zou2024enhancing_aaai,cui_aleth_nerf}. (*: The results of work~\cite{zou2024enhancing_aaai} are directly taken from their paper). Red indicates the best result, while blue indicates the second-best result.}
\label{tab:LOM_low_full}
\Large
\centering
\renewcommand\arraystretch{1.5}
\begin{adjustbox}{max width = 1.00\linewidth}

\begin{tabular}{ccccccc}
\toprule
\toprule
\multicolumn{1}{c|}{\multirow{2}{*}{Method}} & \multicolumn{1}{c|}{\textbf{\textit{``buu"}}}                 & \multicolumn{1}{c|}{\textbf{\textit{``chair"}}}               & \multicolumn{1}{c|}{\textbf{\textit{``sofa"}}}                 & \multicolumn{1}{c|}{\textbf{\textit{``bike"}}}                & \multicolumn{1}{c|}{\textbf{\textit{``shrub"}}}               & \textbf{\textit{mean}}                 \\ \cline{2-7} 
\multicolumn{1}{c|}{}                        & \multicolumn{1}{c|}{PSNR/ SSIM/ LPIPS}   & \multicolumn{1}{c|}{PSNR/ SSIM/ LPIPS}   & \multicolumn{1}{c|}{PSNR/ SSIM/ LPIPS}    & \multicolumn{1}{c|}{PSNR/ SSIM/ LPIPS}   & \multicolumn{1}{c|}{PSNR/ SSIM/ LPIPS}   & PSNR/ SSIM/ LPIPS    \\ \hline
\multicolumn{1}{c|}{3DGS~\cite{3dgs}}   & \multicolumn{1}{c|}{7.53/ 0.299/ 0.442}  & \multicolumn{1}{c|}{6.06/ 0.151/ 0.742}  & \multicolumn{1}{c|}{6.31/ 0.216/ 0.723}  & \multicolumn{1}{c|}{6.37/ 0.077/ 0.781} & \multicolumn{1}{c|}{8.15/ 0.044/ 0.620} &  6.88/ 0.157/ 0.662  \\ \hline
\multicolumn{7}{c}{Image Enhancement Methods + 3DGS}               \\ \hline
\multicolumn{1}{c|}{3DGS + Z-DCE~\cite{Zero-DCE}}         & \multicolumn{1}{c|}{18.02/ 0.834/ 0.303} & \multicolumn{1}{c|}{12.55/ 0.725/ 0.478}  & \multicolumn{1}{c|}{14.66/ 0.822/ 0.460}  & \multicolumn{1}{c|}{10.26/ 0.509/ 0.491}  & \multicolumn{1}{c|}{12.93/ 0.468/ 0.309}  & 13.64/ 0.672/ 0.408 \\
\multicolumn{1}{c|}{Z-DCE~\cite{Zero-DCE} + 3DGS}         & \multicolumn{1}{c|}{17.83/ \blue{0.874}/ 0.350} & \multicolumn{1}{c|}{12.47/ 0.762/ 0.399}   & \multicolumn{1}{c|}{13.86/ 0.841/ \blue{0.308}} & \multicolumn{1}{c|}{10.37/ 0.544/ 0.441} & \multicolumn{1}{c|}{12.74/ 0.487/ \blue{0.248}} &   13.45/ 0.702/ 0.349 \\
\multicolumn{1}{c|}{3DGS + SCI~\cite{cvpr22_sci}}     & \multicolumn{1}{c|}{13.80/ 0.845/ 0.339}                    & \multicolumn{1}{c|}{19.70/ 0.812/ 0.455}  & \multicolumn{1}{c|}{\blue{19.63}/ 0.851/ 0.455}    & \multicolumn{1}{c|}{12.86/ 0.621/0.463}   & \multicolumn{1}{c|}{16.14/ 0.600/ 0.442}   & \multicolumn{1}{c}{15.22/ 0.748/ 0.430} \\
\multicolumn{1}{c|}{SCI~\cite{cvpr22_sci} + 3DGS}              & \multicolumn{1}{c|}{7.68/ 0.690/ 0.523}  & \multicolumn{1}{c|}{11.69/ 0.794/ 0.419}  & \multicolumn{1}{c|}{10.02/ 0.770/ 0.365}  & \multicolumn{1}{c|}{13.55/ 0.667/ \blue{0.390}} & \multicolumn{1}{c|}{15.72/ 0.538/ 0.339}  & \multicolumn{1}{c}{11.73/ 0.692/ 0.407} \\
\multicolumn{1}{c|}{3DGS + NeRCo~\cite{ICCV2023_NeRCo}}            & \multicolumn{1}{c|}{16.64/ 0.765/ 0.401}      & \multicolumn{1}{c|}{19.24/ 0.759/ 0.466}   & \multicolumn{1}{c|}{16.77/ 0.834/ 0.399}                     & \multicolumn{1}{c|}{16.33/ 0.700/ 0.427}                    & \multicolumn{1}{c|}{17.07/ 0.503/ 0.411}                    &       17.21/ 0.712/ 0.421  \\
\multicolumn{1}{c|}{NeRCo~\cite{ICCV2023_NeRCo} + 3DGS}            & \multicolumn{1}{c|}{16.69/ 0.802/ 0.330}     & \multicolumn{1}{c|}{19.11/ 0.773/ 0.376}  & \multicolumn{1}{c|}{18.04/ \blue{0.868}/ 0.381}   & \multicolumn{1}{c|}{16.16/ 0.703/ 0.397}  & \multicolumn{1}{c|}{\blue{17.97}/ 0.502/ 0.399}                    &  17.59/ 0.727/ \blue{0.345}  \\ \hline
\multicolumn{7}{c}{Video Enhancement Methods + 3DGS}                                                               \\ \hline
\multicolumn{1}{c|}{LLVE~\cite{LLVE_2021_CVPR} + 3DGS}             & \multicolumn{1}{c|}{19.67/ 0.868/ \blue{0.253}} & \multicolumn{1}{c|}{15.29/ 0.805/ 0.424} & \multicolumn{1}{c|}{17.18/ 0.858/ 0.379} & \multicolumn{1}{c|}{14.01/ 0.677/ 0.452} & \multicolumn{1}{c|}{15.98/ 0.430/ 0.488} &   16.43/ 0.728/ 0.399 \\
\multicolumn{1}{c|}{SGZ~\cite{SGZ_wacv2022} + 3DGS}    & \multicolumn{1}{c|}{19.21/ 0.832/ 0.270}    & \multicolumn{1}{c|}{12.30/ 0.755/ \blue{0.377}}                & \multicolumn{1}{c|}{14.54/ 0.815/ 0.329}  & \multicolumn{1}{c|}{10.61/ 0.563/ \red{0.375}}     & \multicolumn{1}{c|}{14.04/ \blue{0.565}/ 0.416}   &   14.14/ 0.706/ 0.353  \\ \hline
\multicolumn{7}{c}{NeRF-based Enhancement Methods}                                                                              \\ \hline
\multicolumn{1}{c|}{AME-NeRF*~\cite{zou2024enhancing_aaai}}               & \multicolumn{1}{c|}{\blue{19.89}/ 0.854/ 0.312} & \multicolumn{1}{c|}{17.05/ 0.751/ 0.381} & \multicolumn{1}{c|}{17.93/ 0.847/ 0.378}  & \multicolumn{1}{c|}{18.14/ \blue{0.732}/ 0.437} & \multicolumn{1}{c|}{15.23/ 0.462/ 0.518} & 17.65/ 0.729/ 0.405  \\
\multicolumn{1}{c|}{Aleth-NeRF~\cite{cui_aleth_nerf}(0.45)}       & \multicolumn{1}{c|}{\red{20.22}/ 0.859/ 0.315} & \multicolumn{1}{c|}{\red{20.93}/ \blue{0.818}/ 0.468} & \multicolumn{1}{c|}{19.52/ 0.857/ 0.354}   & \multicolumn{1}{c|}{\red{20.46}/ 0.727/ 0.499} & \multicolumn{1}{c|}{\red{18.24}/ 0.511/ 0.448} & \red{19.87}/ \blue{0.754}/ 0.417  \\
\multicolumn{1}{c|}{Aleth-NeRF~\cite{cui_aleth_nerf}(0.4)}        & \multicolumn{1}{c|}{19.14/ 0.839/ 0.306} & \multicolumn{1}{c|}{16.96/ 0.793/ 0.483} & \multicolumn{1}{c|}{16.97/ 0.847/ 0.367}  & \multicolumn{1}{c|}{17.56/ 0.719/ 0.468} & \multicolumn{1}{c|}{17.55/ 0.484/ 0.451} & \multicolumn{1}{c}{17.64/ 0.736/ 0.415}  \\ \hline
\multicolumn{7}{c}{Our Proposed Method}  
\\ \hline

\multicolumn{1}{c|}{\textbf{Luminance-GS}}            & \multicolumn{1}{c|}{18.09/ \red{0.877}/ \red{0.193}}                    & \multicolumn{1}{c|}{\blue{19.82}/ \red{0.835}/ \red{0.367}}                    & \multicolumn{1}{c|}{\red{20.12}/ \red{0.871}/ \red{0.259}} & \multicolumn{1}{c|}{\blue{18.27}/ \red{0.749}/ 0.411}   & \multicolumn{1}{c|}{15.40/ \red{0.666}/ \red{0.241}} &   \blue{18.34}/ \red{0.799}/ \red{0.294}    \\ 
\bottomrule
\bottomrule
\end{tabular}
\end{adjustbox}
\end{table*}

\begin{table*}[t]
\caption{Per-scene experimental results (PSNR $\uparrow$, SSIM $\uparrow$, LPIPS $\downarrow$) on LOM dataset~\cite{cui_aleth_nerf} overexposure scene, we compare with exposure correction methods~\cite{Afifi_2021_CVPR,Cui_2022_BMVC,zhou2024mslt} and NeRF-based methods~\cite{cui_aleth_nerf}. Red indicates the best result, while blue indicates the second-best result.}.
\label{tab:LOM_oe_full}
\Large
\centering
\renewcommand\arraystretch{1.5}
\begin{adjustbox}{max width = 1\linewidth}

\begin{tabular}{ccccccc}
\toprule
\toprule
\multicolumn{1}{c|}{\multirow{2}{*}{Method}} & \multicolumn{1}{c|}{\textbf{\textit{``buu"}}}                 & \multicolumn{1}{c|}{\textbf{\textit{``chair"}}}               & \multicolumn{1}{c|}{\textbf{\textit{``sofa"}}}                 & \multicolumn{1}{c|}{\textbf{\textit{``bike"}}}                & \multicolumn{1}{c|}{\textbf{\textit{``shrub"}}}               & \textbf{\textit{mean}}                 \\ \cline{2-7} 
\multicolumn{1}{c|}{}                        & \multicolumn{1}{c|}{PSNR/ SSIM/ LPIPS}   & \multicolumn{1}{c|}{PSNR/ SSIM/ LPIPS}   & \multicolumn{1}{c|}{PSNR/ SSIM/ LPIPS}    & \multicolumn{1}{c|}{PSNR/ SSIM/ LPIPS}   & \multicolumn{1}{c|}{PSNR/ SSIM/ LPIPS}   & PSNR/ SSIM/ LPIPS    \\ \hline
\multicolumn{1}{c|}{3DGS~\cite{3dgs}}   & \multicolumn{1}{c|}{6.96/ 0.674/ 0.609}  & \multicolumn{1}{c|}{11.14/ 0.790/ 0.362}  & \multicolumn{1}{c|}{10.17/ 0.790/ 0.369}  & \multicolumn{1}{c|}{9.58/ 0.730/ 0.323} & \multicolumn{1}{c|}{10.34/ 0.646/ 0.299} &   9.64/ 0.726/ 0.392  \\ \hline
\multicolumn{7}{c}{Exposure Correction Methods + 3DGS}               \\ \hline
\multicolumn{1}{c|}{3DGS + MSEC~\cite{Afifi_2021_CVPR}}         & \multicolumn{1}{c|}{16.03/ \blue{0.806}/ 0.517} & \multicolumn{1}{c|}{20.81/ \blue{0.851}/ 0.408}  & \multicolumn{1}{c|}{20.65/ 0.862/ 0.397}  & \multicolumn{1}{c|}{22.10/ 0.826/ 0.305}  & \multicolumn{1}{c|}{18.21/ 0.678/ 0.289}  &  19.56/ 0.805/ 0.382  \\
\multicolumn{1}{c|}{MSEC~\cite{Afifi_2021_CVPR} + 3DGS}         & \multicolumn{1}{c|}{15.08/ 0.804/ 0.440} & \multicolumn{1}{c|}{16.63/ 0.797/ 0.416}   & \multicolumn{1}{c|}{20.09/ 0.828/ 0.335} & \multicolumn{1}{c|}{17.57/ 0.739/ 0.368} & \multicolumn{1}{c|}{16.61/ 0.666/ 0.255} &  17.20/ 0.767/ 0.363  \\
\multicolumn{1}{c|}{3DGS + IAT~\cite{Cui_2022_BMVC}}    & \multicolumn{1}{c|}{15.34/ 0.804/ 0.522}    & \multicolumn{1}{c|}{\blue{21.96}/ 0.833/ 0.292}                    & \multicolumn{1}{c|}{20.23/ \blue{0.872}/ 0.402}                     & \multicolumn{1}{c|}{22.36/ 0.832/ 0.291}                    & \multicolumn{1}{c|}{\red{21.24}/ 0.765/ 0.226}                    & \multicolumn{1}{c}{20.23/ 0.821/ 0.347} \\
\multicolumn{1}{c|}{IAT~\cite{Cui_2022_BMVC} + 3DGS}  & \multicolumn{1}{c|}{15.86/ 0.803/ 0.387}  & \multicolumn{1}{c|}{18.61/ 0.830/ 0.367}  & \multicolumn{1}{c|}{17.42/ 0.833/ 0.348} & \multicolumn{1}{c|}{19.17/ 0.801/ \blue{0.235}}  & \multicolumn{1}{c|}{16.74/ 0.731/ 0.219} & 17.56/ 0.800/ 0.311 \\
\multicolumn{1}{c|}{3DGS + MSLT~\cite{zhou2024mslt}}            & \multicolumn{1}{c|}{15.34/ 0.798/ 0.473}    & \multicolumn{1}{c|}{21.69/ 0.823/ 0.304}                    & \multicolumn{1}{c|}{\red{23.05}/ 0.830/ 0.317}                     & \multicolumn{1}{c|}{23.37/ 0.830/ 0.317}                    & \multicolumn{1}{c|}{\blue{18.89}/ 0.779/ 0.214}                    &   \multicolumn{1}{c}{\blue{20.39}/ 0.815/ 0.345}  \\
\multicolumn{1}{c|}{MSLT~\cite{zhou2024mslt} + 3DGS}            & \multicolumn{1}{c|}{16.35/ 0.805/ \blue{0.333}}     & \multicolumn{1}{c|}{20.93/ 0.828/ \blue{0.275}}  & \multicolumn{1}{c|}{\blue{21.65}/ 0.847/ \blue{0.259}}   & \multicolumn{1}{c|}{\blue{24.03}/ \blue{0.841}/ 0.244}  & \multicolumn{1}{c|}{18.29/ \red{0.797}/ \blue{0.199}}   &  \multicolumn{1}{c}{20.25/ \blue{0.824}/ \blue{0.262}} \\ \hline

\multicolumn{7}{c}{NeRF-based Exposure Correction Method}                                             \\ \hline

\multicolumn{1}{c|}{Aleth-NeRF~\cite{cui_aleth_nerf}}       & \multicolumn{1}{c|}{\blue{16.78}/ 0.805/ 0.611} & \multicolumn{1}{c|}{20.08/ 0.820/ 0.499} & \multicolumn{1}{c|}{17.85/ 0.852/ 0.458}   & \multicolumn{1}{c|}{19.85/ 0.773/ 0.392} & \multicolumn{1}{c|}{15.91/ 0.477/ 0.483} & 18.09/ 0.745/ 0.488  \\
 \hline
\multicolumn{7}{c}{Our Proposed Method}                                                                                                                                                                                                                                                 \\ \hline

\multicolumn{1}{c|}{\textbf{Luminance-GS}}            & \multicolumn{1}{c|}{\red{19.67}/ \red{0.811}/ \red{0.311}}                    & \multicolumn{1}{c|}{\red{22.63}/ \red{0.856}/ \red{0.207}}                    & \multicolumn{1}{c|}{21.16/ \red{0.878}/ \red{0.204}}                     & \multicolumn{1}{c|}{\red{24.05}/ \red{0.851}/ \red{0.216}}                    & \multicolumn{1}{c|}{16.04/ \blue{0.780}/ \red{0.173}}   & \red{20.71}/ \red{0.835}/ \red{0.222}    \\ 
\bottomrule
\bottomrule
\end{tabular}
\end{adjustbox}
\end{table*}

\begin{figure*}[t]
    \centering
    \includegraphics[width=1\linewidth]{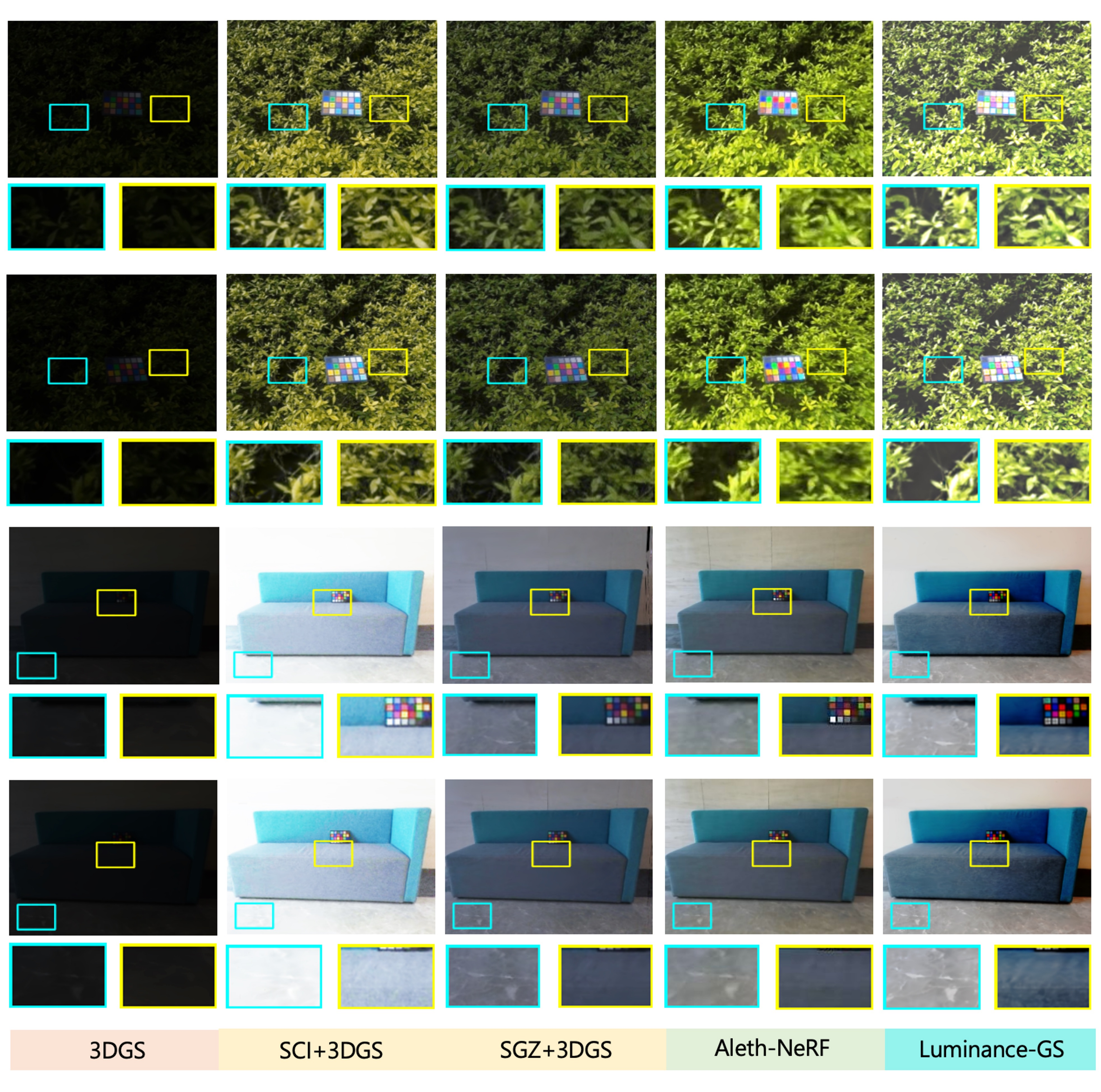}
    
    \caption{Novel view synthesis results in LOM dataset low-light “\textbf{\textit{shrub}}” and “\textbf{\textit{sofa}}” scenes, we show the comparison results with 3DGS~\cite{3dgs}, combination of low-light enhancement methods (SCI~\cite{cvpr22_sci}, SGZ~\cite{SGZ_wacv2022}) with 3DGS and Aleth-NeRF~\cite{cui_aleth_nerf}.}
    \label{fig:supp_low}
    
\end{figure*}

\begin{figure*}[t]
    \centering
    \includegraphics[width=1\linewidth]{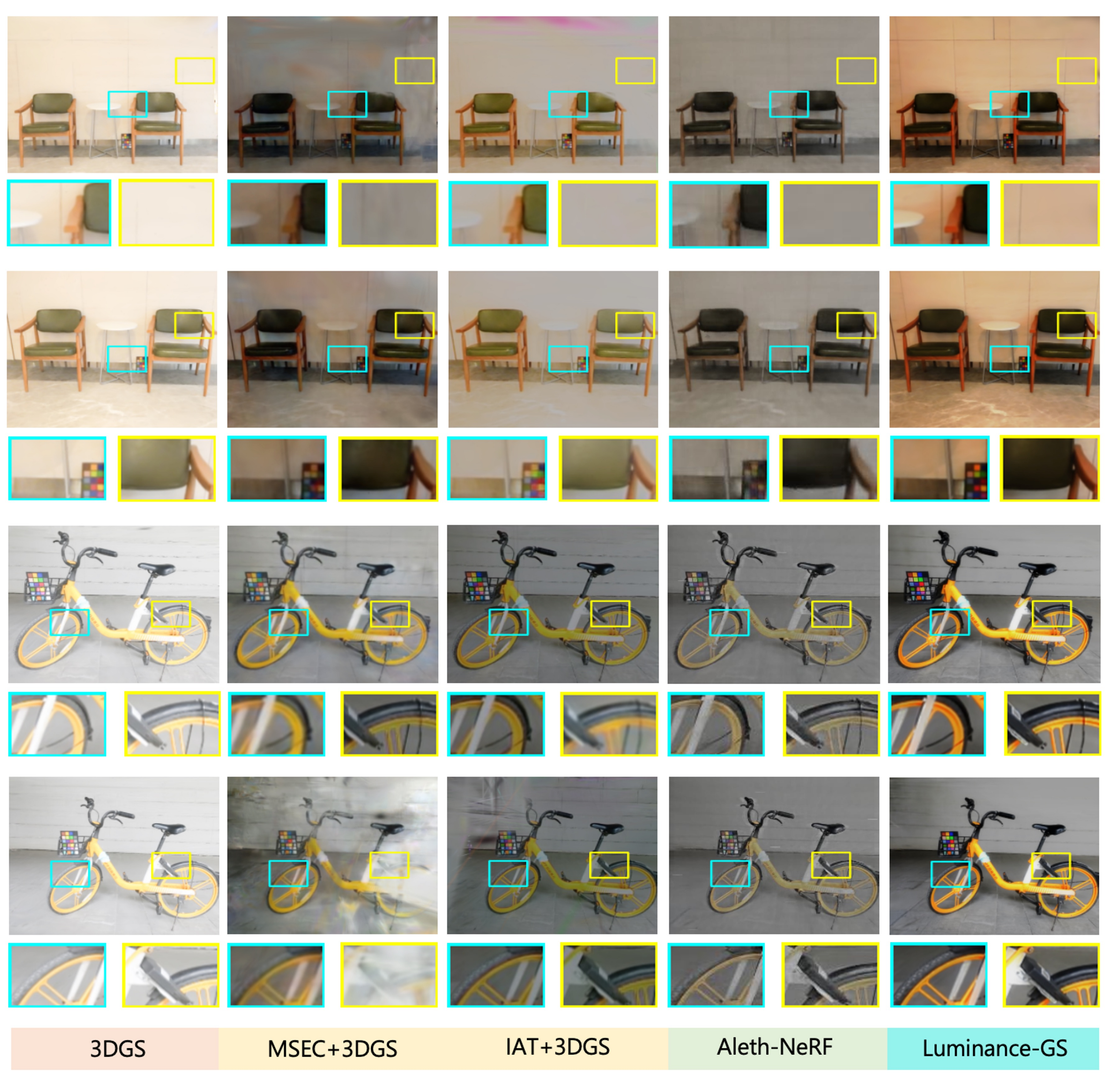}
    
    \caption{Novel view synthesis results in LOM dataset over-exposure “\textbf{\textit{chair}}” and “\textbf{\textit{bike}}” scenes, we show the comparison results with 3DGS~\cite{3dgs}, combination of exposure correction methods (MSEC~\cite{Afifi_2021_CVPR}, IAT~\cite{Cui_2022_BMVC}) with 3DGS and Aleth-NeRF~\cite{cui_aleth_nerf}.}
    \label{fig:supp_exposure}
    
\end{figure*}

\begin{figure*}[t]
    \centering
    \includegraphics[width=1\linewidth]{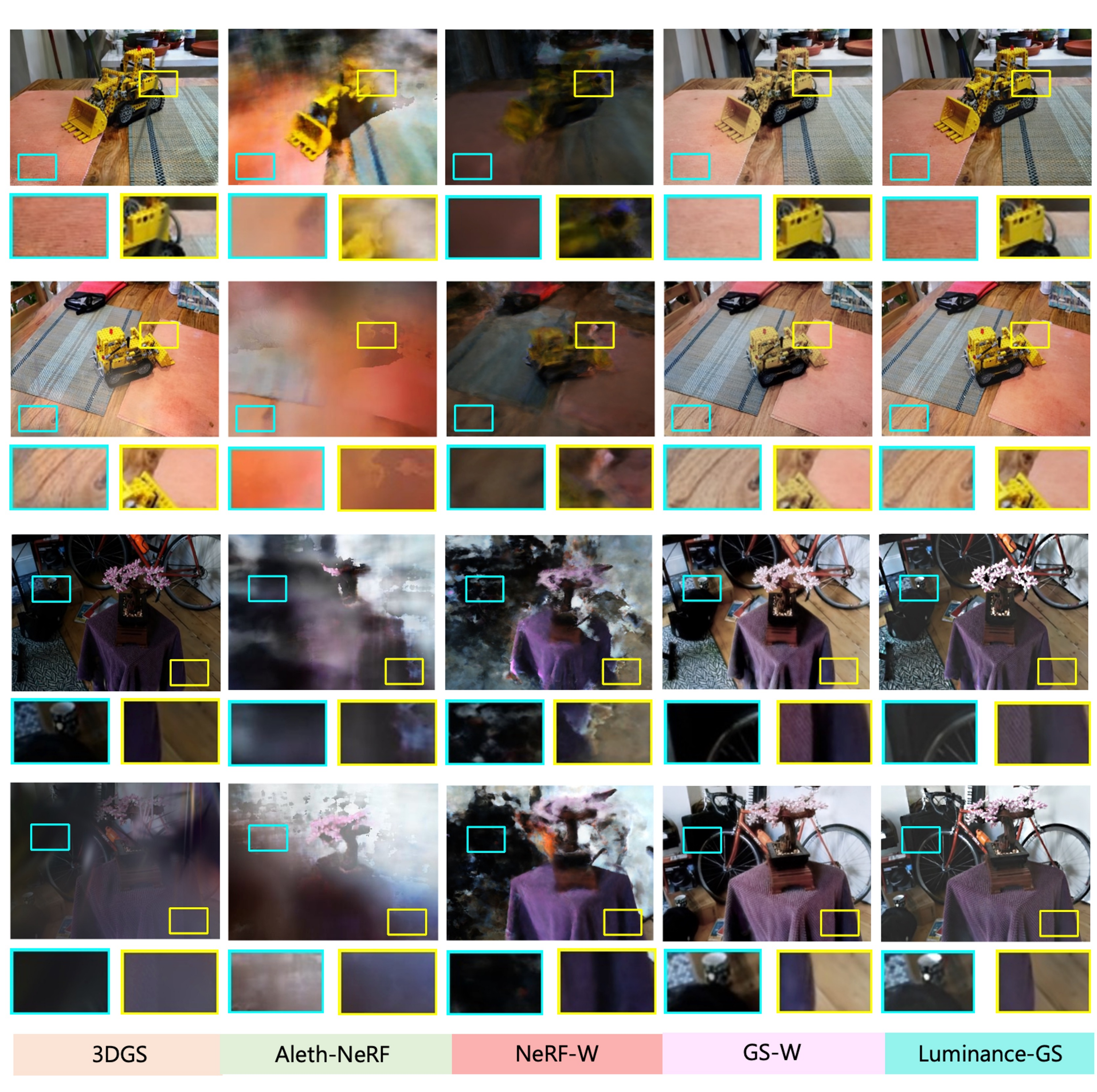}
    
    \caption{Novel view synthesis results on our synthesized varying exposure unbounded dataset (from Mip-NeRF 360 dataset~\cite{barron2021mipnerf}) “\textbf{\textit{kitchen}}” and “\textbf{\textit{bonsai}}” scenes, with comparison of 3DGS~\cite{3dgs}, Aleth-NeRF~\cite{cui_aleth_nerf}, NeRF-W~\cite{martinbrualla2020nerfw} and GS-W~\cite{GS-W_ECCV2024}.}
    \label{fig:supp_uneven}
    
\end{figure*}

\end{document}